\documentclass[final]{cvpr}

\usepackage{times}
\usepackage{epsfig}
\usepackage{graphicx}
\usepackage{amsmath}
\usepackage{amssymb}
\usepackage{subfig}
\usepackage{booktabs}
\usepackage{multirow}
\usepackage{float}

\usepackage[pagebackref=true,breaklinks=true,colorlinks,bookmarks=false]{hyperref}

\def\cvprPaperID{4645} % *** Enter the CVPR Paper ID here
\def\confYear{CVPR 2021}

\begin{document}

%%%%%%%%% TITLE
\title{Learning Category-level Shape Saliency via Deep Implicit Surface Networks}

\author{Chaozheng Wu$^1$, Lin Sun$^2$, Xun Xu$^3$, Kui Jia$^1$\thanks{Corresponding author} \\  
$^1$South China University of Technology, $^2$Samsung, USA, $^3$I2R, ASTAR\\
{\tt\small eeczwu@mail.scut.edu.cn}, {\tt\small sunlin@cs.stanford.edu}, \\
{\tt\small alex.xun.xu@gmail.com}, {\tt\small kuijia@scut.edu.cn}
}

% \and

\maketitle

%%%%%%%%% ABSTRACT
\begin{abstract}
   This paper is motivated from a fundamental curiosity on what defines a category of object shapes. For example, we may have the common knowledge that a plane has wings, and a chair has legs. Given the large shape variations among different instances of a same category, we are formally interested in developing a quantity defined for individual points on a continuous object surface; the quantity specifies how individual surface points contribute to the formation of the shape as the category. We term such a quantity as category-level shape saliency or shape saliency for short. Technically, we propose to learn saliency maps for shape instances of a same category from a deep implicit surface network; sensible saliency scores for sampled points in the implicit surface field are predicted by constraining the capacity of input latent code. We also enhance the saliency prediction with an additional loss of contrastive training. We expect such learned surface maps of shape saliency to have the properties of smoothness, symmetry, and semantic representativeness. We verify these properties by comparing our method with alternative ways of saliency computation. Notably, we show that by leveraging the learned shape saliency, we are able to reconstruct either category-salient or instance-specific parts of object surfaces; semantic representativeness of the learned saliency is also reflected in its efficacy to guide the selection of surface points for better point cloud classification. 
\end{abstract}

%%%%%%%%% BODY TEXT
\section{Introduction}
\begin{figure}[htbp]
\centering
% \subfloat[ISSN]{
% \includegraphics[width=0.6\linewidth]{figures/network1.png} \label{net1}
% \caption{Vanilla}
% }
% \quad
% \subfloat[Implicit surface saliency networks]{
\includegraphics[width=1.0\linewidth]{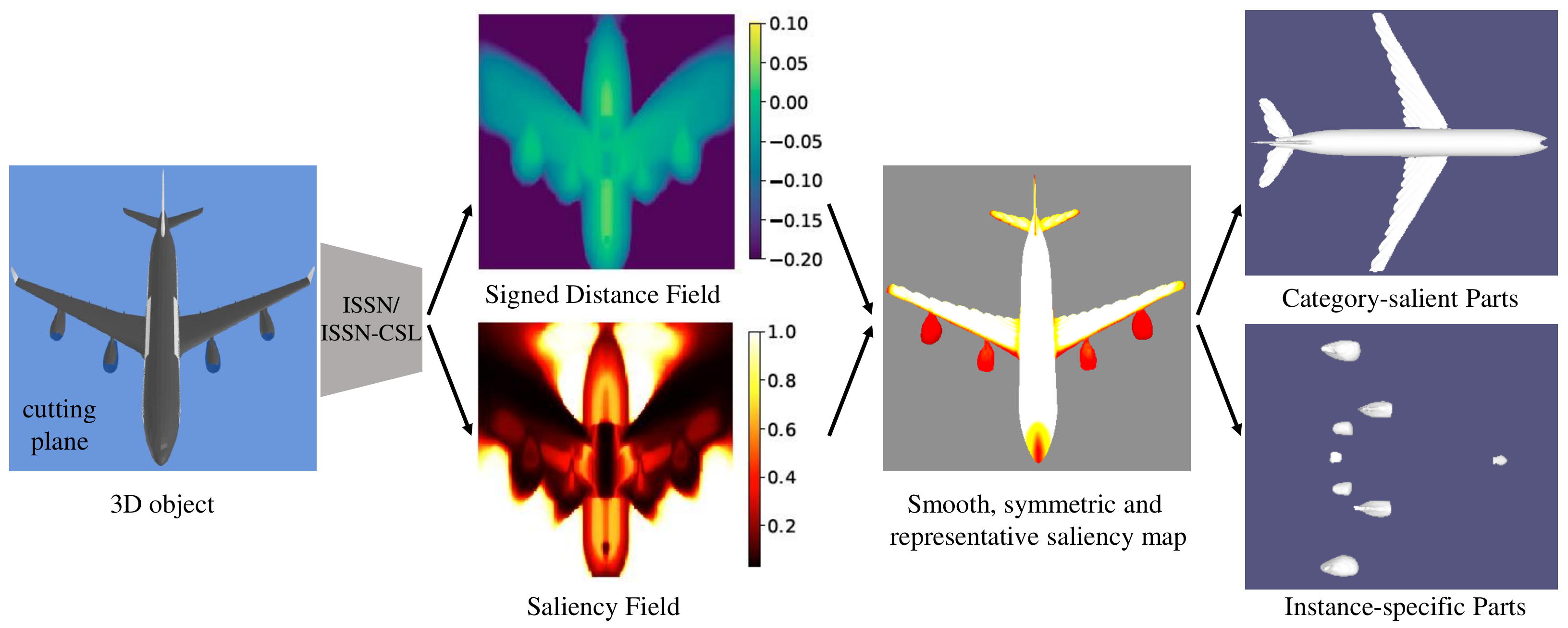} %\label{net2}
% \caption{With contrastive classification network}
% }
\caption{Abstract concept of ISSN and ISSN-CSL, that can learn continuous signed distance fields and saliency fields of objects in 3D space simultaneously. The learned category-level saliency maps have the properties of smoothness, symmetry, and semantic representativeness. By leveraging the learned shape saliency, we are able to reconstruct either category-salient or instance-specific parts of object surfaces. }\label{pull-image}
\vspace{-1.5em}
\end{figure}

The conception of saliency in cognitive science is referred to the information of interest filtered by human visual system (HVS) \cite{berger2015looking} which is distinct from those around it. Saliency learning is important in computer vision, especially in pattern recognition, as it help us understand what contributes to the success of recognition and what features are of high importance with comparison to others \cite{Simonyan2014DeepIC,zheng2019pointcloud}. One uses the knowledge of saliency to localize the target objects in the images \cite{Simonyan2014DeepIC}, design adversarial attack \cite{papernot2016limitations}, \etc. Moreover, saliency map estimation on 3D data can benefit many applications. For example, Lee \etal introduce the mesh saliency and apply it to mesh simplification and good view selection. Song \etal \cite{song2014mesh} estimate mesh saliency via spectral processing and apply it to mesh segmentation and scan integration. Saliency map estimation also shows potentials for other applications such as shape matching and retrieval, alignment and sampling. 

Compared to 2D image processing, saliency map estimation on 3D data is challenging due to following factors: (1) unlike the regular and discrete representation of 2D image, 3D data is irregular and continuous, \eg the locations of the mesh vertices can be anywhere in 3D space, (2) 3D data is invariant to permutation, \eg if we shuffle the order of the point cloud, its appearance does not change, (3) the degree of freedom of transformation in 3D space is more than that in 2D space, making it hard to learn transformation invariant features.

There has been numerous studies to investigate the mesh saliency which can be categorized into local contrast based and global contrast based methods. Many of the local contrast based methods \cite{harris:sipiran2010robust,harris:sipiran2011harris,zaharescu2009surface,sift:godil2011salient} extend the classical operators in 2D image processing to 3D space to detect the local variance, \eg Harris \cite{harris1988combined}, SIFT \cite{sift:lowe2004distinctive}, \etc. Nevertheless, the saliency maps estimated by these methods are dispersive and small-scale. On the other hand, global contrast based methods \cite{global:song2012conditional,global:sipiran2013key,global:wang2015multi} try to estimate saliency maps in large scale. However, most of them are built on local contrast based methods. As a result, the saliency maps estimated by these methods are still lack of representativeness. Very recently, following \cite{Simonyan2014DeepIC}, Zheng \etal \cite{zheng2019pointcloud} introduce the Point Cloud Saliency Map (PCSM) to quantify the contribution of each point to the semantic classification task via the backward gradient on the point. Although such saliency maps can capture the representativeness of the object to some extend, they look noisy and discontinuous which is a common issue of such gradient based methods \cite{gupta20203d}.

Recently, 3D shape reconstruction has been attracting great attention in computer vision, we find that the dominant methods can reconstruct the common parts shared among objects of the same category much easier than the instance-specific details. In this paper, we rethink two fundamental problems in 3D saliency map estimation about what defines a category of object shapes and what contributes to discriminate the shapes of different categories. Naturally, we turn to 3D shape reconstruction to learn the representativeness of a category of objects. Specifically, we choose a recently proposed implicit fields learning framework DeepSDF \cite{park2019deepsdf} as our backbone, where the continuous signed distance function (SDF) is learned to represent the object shape, to learn a continuous saliency maps along the manifolds of the object surfaces. We term our proposed method the Implicit Surface Saliency Network (ISSN). Furthermore, motivated by Tags2Parts \cite{muralikrishnan2018tags2parts} where the region of interest specified by the tags can be detected using a binary classification network, we introduce the contrastive saliency learning to generate the saliency maps that can discriminate between the shapes of the category of interest and those of other categories. We term our implicit surface saliency network with contrastive saliency learning the ISSN-CSL. The abstract concept of ISSN and ISSN-CSL is shown in Figure \ref{pull-image}. Unlike PCSM \cite{zheng2019pointcloud} where saliency maps are obtained by backward gradient of classification network, both our proposed ISSN and ISSN-CSL generate the saliency maps via a feedforward decoder. Experiments show that the saliency maps learned by ISSN and ISSN-CSL have excellent properties of smoothness, symmetry, and semantic representativeness. Notably, by leveraging the learned shape
saliency of ISSN, we are able to reconstruct either category-salient or instance-specific parts of object surfaces. We summarize our contributions as follows:
\begin{itemize}
\item We are the first to introduce two implicit surface saliency network, ISSN and the one with contrastive saliency learning ISSN-CSL, to learn category-level shape saliency via deep implicit surface networks.

\item To compare the smoothness and symmetry of saliency maps of different methods quantitatively, we introduce two evaluation metrics, smoothness ratio and symmetry distance of saliency. To quantify the representativeness of the saliency maps, we design the saliency points classification task.

\item We demonstrate that the saliency maps learned by our proposed ISSN and ISSN-CSL have better properties of smoothness, symmetry, and semantic representativeness than those of other methods. We also conduct experiments to show that the saliency maps of ISSN can be used to reconstruct either category-salient or instance-specific parts of object surfaces.

\end{itemize}

%------------------------------------------------------------------------
\section{Related Works}

\subsection{3D Shape Reconstruction}
Traditionally, people reconstruct the complete 3D object surface by using the stereo correspondences from multi-view 2D images. 3D shape reconstruction from single image using learning-based methods have been explored in prior studies. Most of them decode the embedding of 2D image into 3D voxel \cite{recon:choy20163d,recon:girdhar2016learning,recon:wu2018learning}. However, the reconstructed results of these approaches are low-resolution due the sparsity of 3D voxel and limitation of GPU memory. Most relevant to our work, Chen and Zhang \cite{recon:chen2019learning} propose the IM-Net to learn an implicit field of object in the continuous 3D space which is divided into two parts, inside and outside the object, and the implicit field learning can be recast as a binary classification task. Similarly, Park \etal \cite{park2019deepsdf} propose the DeepSDF to estimate the continuous Signed Distance Function (SDF) in 3D space. For the space inside the object, $SDF<0$, and for the space outside the object, $SDF>0$. The surface of object is implicitly encoded on the manifold where $SDF=0$. 
The major differences between IM-Net and DeepSDF are that IM-Net adopts an auto-encoder network and learns the implicit field via a classification loss, while DeepSDF introduces a novel auto-decoder network to regress the continuous SDF values.
% A major difference between IM-Net and DeepSDF is that IM-Net adopt the auto-encoder network, while DeepSDF introduce a novel auto-decoder network. 
After training the implicit field, both IM-Net \cite{recon:chen2019learning} and DeepSDF \cite{park2019deepsdf} adopt the Marching Cubes algorithm to obtain the object surface and the reconstructed object can be achieved at arbitrary resolution depending on the grid size. 

% \vspace{0.2cm}
\subsection{Saliency Map Estimation}
In the field of computer science, saliency firstly emerged in 2D image analysis. With the increasing demand of geometric analysis and processing on 3D data, saliency map estimation is drawing attention in graphics community recently. Borrowing ideas from 2D image processing, Sipiran and Bustos \cite{harris:sipiran2010robust,harris:sipiran2011harris} introduce the 3D version of Harris operator \cite{harris1988combined}, Zaharescu \etal \cite{zaharescu2009surface} develop the MeshDoG to detect the salient points, and Godil and Wagan \cite{sift:godil2011salient} extend the SIFT operator \cite{sift:lowe2004distinctive} to 3D space. Song \etal \cite{global:song2012conditional} propose the conditional-random field-based (CRF) method to detect mesh saliency, which is more effective and robust. To capture semantic representativeness of point cloud data, Zheng \etal \cite{zheng2019pointcloud} and Gupta \etal \cite{gupta20203d} propose to estimate the point cloud saliency maps using the gradients of trained classification networks. In contrast, we take a rather different approach that learns the continuous shape saliency maps via deep implicit surface networks.

% \vspace{0.2cm}
\subsection{Contrastive Learning}
With the rapid development of self-supervised learning, contrastive learning is getting more and more attention. Wu \etal \cite{cl:wu2018unsupervised} treat each image instance as a category and utilize noise-contrastive estimation (NCE) \cite{gutmann2010noise} to tackle challenge of computing the similarity among a large number of instances, aiming to capture the instance-level apparent similarity. Tian \etal \cite{cl:tian2019contrastive} propose the contrastive multi-view coding to learn the features shared among different views of the same scene. Misra and Maaten \cite{cl:misra2020self} introduce the Pretext-Invariant Representation Learning (PIRL) that learns representations invariant to transformation by solving the jigsaw puzzles. Chen \etal \cite{cl:chen2020simple} design the simplified contrastive learning of visual representations (SimCLR) that does not require memory bank or special architectures. He \etal \cite{cl:he2020momentum} develop the Momentum Contrast (MoCo) where they build a dynamic dictionary with a queue and a moving-averaged encoder so as to accelerate contrastive learning. Grill \etal \cite{cl:grill2020bootstrap} propose an approach BYOL that does not require negative samples and replace the NCE-based loss with L2 loss. In 3D space, Xie \etal \cite{cl:xie2020pointcontrast} propose PointContrast to learn representations on 3D real point cloud data. They find that the pretrained model can be used to boost the performances of existing methods on several benchmarks of 3D segmentation and detection. Muralikrishnan \etal \cite{muralikrishnan2018tags2parts} utilize a simple binary classification network to specify whether the input object contains the region specified by the tag or not. As a result, they can segment the region of interest from other regions on the object. Similarly, we propose the contrastive saliency learning based on a binary classification network to capture the representativeness of the category of interest.
%------------------------------------------------------------------------
\section{The Proposed Method}
We illustrate two methods, Implicit Surface Saliency Networks (ISSN) and Implicit Surface Saliency Networks with Contrastive Saliency Learning (ISSN-CSL), to estimate the shape saliency maps for objects from the same category via deep implicit surface networks which are initially designed for shape reconstruction. Our motivation lies in that the intra-category objects share some common parts and it's easy for the network to estimate the implicit field of these common parts, while it may not be easy to estimate the implicit field of the instance-specific parts. Therefore, we design the network to output saliency score associated with the SDF value for every query point, and use the saliency score to weight the corresponding SDF error (Section \ref{issn}). The saliency map can represent the category commonality of the object surface. 
% We call such inductive saliency map estimation method the ISSN. 
However, this kind of saliency map may not capture the semantic representativeness of the object. In order to generate the saliency map that can capture the semantic cues for the objects from the category of interest, we introduce the contrastive saliency learning that discriminates between instances of the category of interest and those of other categories (Section \ref{issn-csl}). 
% We call it the transductive saliency map estimation (ISSN-CSL).
% The overall architecture is shown in Fig. \ref{overall}. Finally, we present the applications of saliency map for downstream tasks (\eg point cloud classification, part segmentation, \etal).

\begin{figure}[htbp]
\centering
% \subfloat[ISSN]{
% \includegraphics[width=0.6\linewidth]{figures/network1.png} \label{net1}
% \caption{Vanilla}
% }
% \quad
% \subfloat[Implicit surface saliency networks]{
\includegraphics[width=1.0\linewidth]{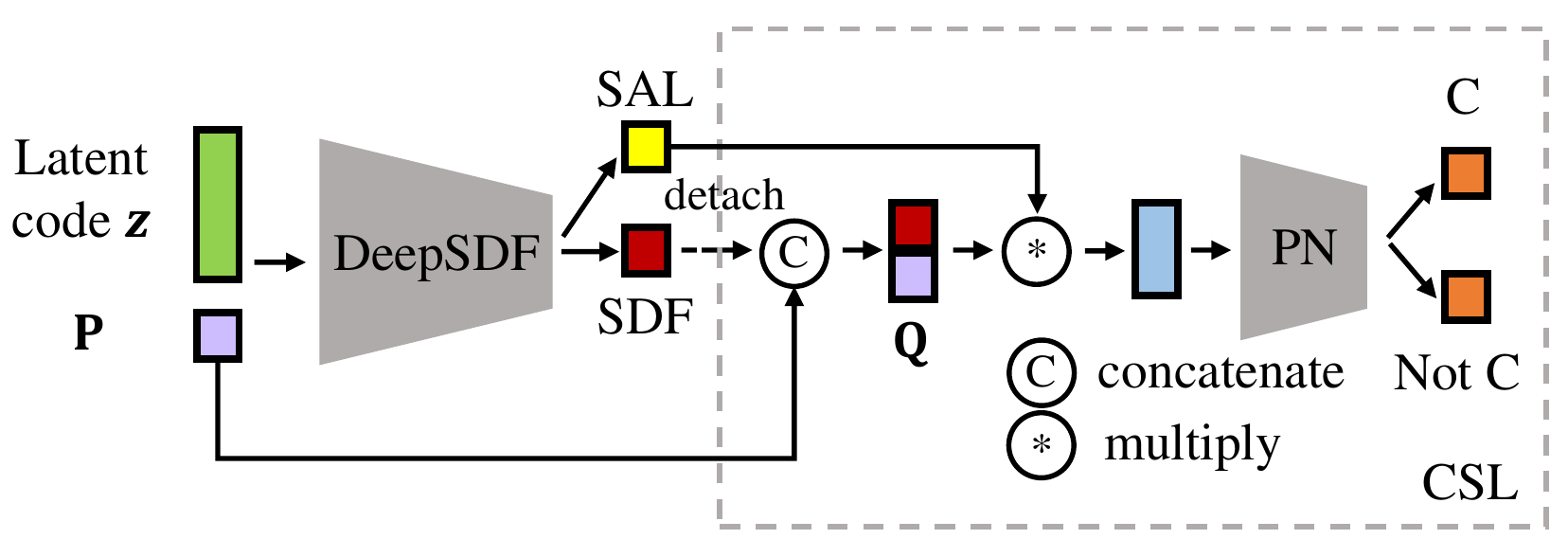} %\label{net2}
% \caption{With contrastive classification network}
% }
\caption{The Architecture of Proposed ISSN and ISSN-CSL. The input of the networks are the latent code $\mathbf{z}$ and a set of query points $\mathbf{P}$, and the output are the SDF values and saliency scores (SAL). For contrastive saliency learning (CSL) , the points in $\mathbf{P}$ are concatenated with their predicted SDF values (denoted as $\mathbf{Q}$) and multiplied by the saliency scores, then pass through a PointNet (PN) \cite{qi2017pointnet} to specify whether the object denoted by $\mathbf{Q}$ is from the category of interest (C) or not.}\label{network}
\vspace{-0.8em}
\end{figure}

\subsection{Implicit Surface Saliency Network} \label{issn}
The architecture of ISSN is illustrated in Figure \ref{network}. It is built on a Sign Distance Function (SDF) estimation network \cite{park2019deepsdf}. The input of the network are the latent code $\mathbf{z}$ that encodes the shape of object and query points $\mathbf{P}$ in 3D space. Different from DeepSDF, our network outputs the point-wise SDF values and saliency scores (SAL) simultaneously. To make the network predict the saliency scores in an unsupervised manner, we apply the loss function:
\begin{equation}\label{loss:issn}
\mathcal{L}_{issn} = \frac{1}{N}\sum_{i=0}^{N} (\mathcal{L}_{i}s_i-\lambda\log(s_i)) + \beta\lVert\mathbf{z}\rVert_2^2,
\end{equation}
where $N$ denotes the number of input query points, and $s_i$ is the saliency score of input query point $\mathbf{p}_i$. Similar to \cite{park2019deepsdf}, $\mathcal{L}_i$ is the $L_1$ loss that penalizes $L_1$ distance between the predicted and Ground-Truth (GT) SDF values,

\begin{equation}\label{loss:sdf}
\mathcal{L}_{i} = \left| clamp(f_i, \delta) - clamp(\hat{f}_i, \delta)\right|,
\end{equation}
where $f_i$ and $\hat{f}_i$ are the predicted and GT SDF values of query point $\mathbf{p}_i$. Please refer to \cite{park2019deepsdf} for the definition of $clamp$ function and the choice of $\delta$. We introduce a regularization term $\log(s_i)$ in Equation (\ref{loss:issn}) to avoid vanishing of $s_i$, and $\lambda$ is to balance the weighted SDF loss and saliency scores.
% The same strategy is also used in \cite{wang2019densefusion}. 
Besides, we add $L2$-norm of the latent $\mathbf{z}$ to constrain the expressiveness so that decoder can focus on the reconstruction of intra-category common part while omits the instance-specific details. Note that if $\lambda$ in Equation (\ref{loss:issn}) is set to $0$, we will not blend $s_i$ with $L_i$, the loss will degrade to DeepSDF loss. 

To understand why the proposed method works, here, we provide more insights into it. An object instance consists of common parts $\mathbb{C}$ shared among objects from the same category and instance-specific details $\mathbb{D}$. It is easier for the network to regress the SDF values of query points around $\mathbb{C}$ than those around $\mathbb{D}$. Thus, for $\mathbf{p}_i$ around $\mathbb{C}$, $\mathcal{L}_i$ will have smaller value, then incur high penalty from regularization term $-\log(s_i)$ and lead to a high value of $s_i$. On the contrary, for $\mathbf{p}_i$ around $\mathbb{D}$, $\mathcal{L}_i$ will have larger value, then bring about low value of $s_i$.

An alternative way to estimate this kind of saliency map is using the Principle Component Analysis (PCA). As shown in \cite{pauly2003multi,bae2008method}, PCA is usually used to estimate the curvature of the manifold, which captures the the local variance of single object, while ours is able to capture how individual surface points contribute to the formation of the shape as the category.

\subsection{Contrastive Saliency Learning} {\label{issn-csl}}

% Besides the inductive saliency map estimation method as mentioned in Section \ref{issn-in}, there are some transductive saliency map estimation methods, \eg PointCloud Saliency Map (PCSM) \cite{zheng2019pointcloud}. 
The design strategy in Section \ref{issn} enables the network to capture the common parts shared among the objects of the same category, but it may be lack of semantic representativeness which discriminates between instances of the category of interest and those of other categories.
In PCSM \cite{zheng2019pointcloud}, the authors proposed to estimate the point cloud saliency map based on the derivative of the loss function of a trained classification network with respect to the input point cloud. Although the saliency map of PCSM contains the semantic information to some extent, PCSM inevitably face the noise and discontinuities which are common in gradient based saliency map \cite{gupta20203d}. 
%and our experiment also shows that the saliency map of PCSM looks noisy and not smooth. 
Inspired by the Tags2Parts \cite{muralikrishnan2018tags2parts} where the region of interest specified by the tag can be segmented via a simple binary classification networks, we introduce a contrastive classification network following the ISSN to estimate the saliency map that can capture the semantic cues of the objects from the category of interest, giving rise to ISSN-CSL.

During training, we sample the instances in $C$ and those not in $C$ equally in a mini batch. Besides the ISSN regression loss in Equation (\ref{loss:sdf}), we add a binary classification loss:
\begin{equation}
\mathcal{L}_{cls} = -\hat{l}\log(l) - (1-\hat{l})\log(1-l),
\end{equation}
where $\hat{l}$ and $l$ are the GT label and predicted score, respectively. For the shapes of the category of interest, $\hat{l}=1$, otherwise, $\hat{l}=0$. The total loss function for ISSN-CSL is:
% \begin{equation}
% \mathcal{L}_{csl} = \frac{1}{N}\sum_{i=0}^{N} \mathcal{L}_{i} + \gamma \mathcal{L}_{con} + \beta\lVert\mathbf{z}\rVert_2^2.
% \end{equation}
\begin{equation}
\mathcal{L}_{csl} = \mathcal{L}_{issn} + \gamma \mathcal{L}_{cls},
\end{equation}
where $\gamma$ is a hyperparameter to balance $\mathcal{L}_{issn}$ and $\mathcal{L}_{cls}$.

% Besides the intra-category saliency map that captures the common parts shared between the objects from the same category, we also introduce the inter-category saliency map that captures the discriminative parts of the category of interest from other categories. To generate the inter-category saliency map, we introduce a contrastive classification network (CCN) as shown in Fig. \ref{net2}. 
At last, we review the architecture step by step, as shown in Figure \ref{network}, given a point set $\mathbf{P}$ with $N$ points and the latent code $\mathbf{z}$ with $D$ dimension, we firstly duplicate the latent code $N$ times and concatenate it with the point coordinates in $\mathbf{P}$, then feed them to the DeepSDF network to predict the point-wise SDF values $\mathbf{F}$ and saliency scores $\mathbf{S}$. After that, $\mathbf{F}$ is detached from DeepSDF network and concatenated with $\mathbf{P}$, formed $\mathbf{Q}$. $\mathbf{Q}$ can represent the shape of the object. Finally, we blend $\mathbf{S}$ with $\mathbf{Q}$ and feed it into a PointNet \cite{qi2017pointnet} to classify whether $\mathbf{Q}$ is from the category of interest $C$ or not. For those points around the discriminative parts of the category of interest, they will contribute more to the contrastive classification, then the corresponding saliency scores will be higher. We will show the saliency map of ISSN-CSL has the potential to boost the performance of point cloud classification in the later part.

\subsection{Saliency Map Properties }\label{analysis}
To evaluate the properties of smoothness and symmetry of the saliency maps estimated quantitatively, we introduce two evaluation metrics, saliency smoothness ratio and saliency symmetry distance. 
% Furthermore, we evaluate the representativeness and discrimination of saliency maps by developing the saliency points cluttering and classification.
% The saliency roughness is to evaluate the smoothness of the saliency map and the saliency symmetry distance is to evaluate the symmetry property of saliency maps for the symmetric objects.

% \subsubsection{Physical Analysis}

% \subsubsection{Reconstruction Analysis}
% As shown in Figure \ref{fig:recon}.

\subsubsection{Saliency Smoothness}\label{smooth}
\
% The saliency roughness is to evaluate the roughness of the saliency map, the smaller value denotes that the saliency map is more smoother.
%\quad Given the saliency maps $\mathbf{S}$ of a point cloud $\mathbf{P}$ generated by our proposed methods or other methods \cite{zheng2019pointcloud,bae2008method}, we analyze their smoothness qualitatively and quantitatively. In Figure \ref{fig:pc_sal}, we can find that the saliency map generated by our proposed method looks much smoother than that of PCSM \cite{zheng2019pointcloud} and PCA \cite{bae2008method}. For quantitative comparison, we define the saliency smoothness ratio ($SR$) similar to contrast ratio as following:
\quad Given the saliency maps $\mathbf{S}$ of a point cloud $\mathbf{P}$, to quantitatively evaluate the saliency smoothness, we define our saliency smoothness ratio (SSR) as follows,

\begin{equation}
SSR = \frac{1}{N}\sum_{i=1}^N \left|s_i - \tilde{s_i}\right|,
\end{equation}
where $\tilde{s_i}$ is the saliency score of the nearest neighbor of point $\mathbf{p}_i$. 
% For fair comparison, the saliency map of a point cloud is normalized to $[0,1]$. 
% The saliency smoothness is to evaluate the smoothness of the saliency map, the smaller value indicates that the smoother saliency map.
The smaller value of $SSR$ indicates the smoother saliency map.

\subsubsection{Saliency Symmetry}\label{symmetry}

\quad We denote the point cloud of an object by $\mathbf{P}$ and obtain its mirror point cloud $\mathbf{P}^{'}$ by flipping $\mathbf{P}$ along the plane of symmetry. The plane of symmetry for each category need to be searched manually. For each $\mathbf{p}_i^{'}\in\mathbf{P}^{'}$, we find its nearest neighbor $\mathbf{p}_i$ on $\mathbf{P}$, as illustrated in Figure \ref{fig:symmetry}. If the maximum distance between $\mathbf{p}_i^{'}$ and $\mathbf{p}_i$ is less than $10cm$, we consider it a symmetry object, otherwise, it is asymmetric.

% For the symmetric objects, we can also conduct the symmetry analysis. If $\mathbf{P}$ is the point cloud from a symmetric object, for each $\mathbf{p}_i\in\mathbf{P}$, we find its nearest mirror point $\mathbf{p}_i^{'}$ about the symmetry axis, as illustrated in Figure \ref{fig:symmetry}. 
To evaluate the symmetry of saliency map quantitatively, we define the symmetry distance of saliency as following:
\begin{equation}
d_{sym} = \frac{1}{N}\sum_{i=1}^N \left|s_i - s_i^{'}\right|,
\end{equation}
where $s_i$ and $s_i^{'}$ are the saliency scores of $\mathbf{p}_i$ and $\mathbf{p}_i^{'}$, respectively. 
% For fair comparison, the saliency map of a point cloud is normalized to [0,1].

% \begin{figure}[ht]
% \centering
% \includegraphics[width=0.7\linewidth]{figures/symmetry.png}
% \caption{Symmetry analysis.}\label{fig:symmetry}
% \end{figure}
% Add to Figure 2

% \subsection{Object Part Segmentation} {\label{sal_seg}}
% % \noindent\textbf{Plan A} - 
% For the sampled point cloud, we feed it into the trained saliency network to predict the point-wise saliency scores. Then we use the saliency score to weight the point-wise classification loss in part segmentation. We denote by $l(\mathbf{p}_i)$ and $\hat{l}(\mathbf{p}_i)$ the ground truth labels and predicted score of $\mathbf{p}_i$, respectively. Then the segmentation loss can be written as following:
% \begin{equation}
% \mathcal{L}_{seg} = \frac{1}{N}\sum_{i=1}^N -\hat{l}(\mathbf{p}_i)\log(\hat{l}(\mathbf{p}_i))-(1-\hat{l}(\mathbf{p}_i))\log(1-\hat{l}(\mathbf{p}_i)).
% \end{equation}

% \noindent\textbf{Plan B} - If \textbf{Plan A} does not work, maybe we should try to input the saliency map along with the point cloud to the segmentation network at the beginning.

%------------------------------------------------------------------------
\section{Experiments}
\subsection{Datasets}
To verify our proposed method, two public datasets, ModelNet40 \cite{wu20153d} and ShapeNet \cite{chang2015shapenet} are used. The objects in both datasets are aligned to canonical poses.

\vspace{0.2cm}
\noindent\textbf{ModelNet40} \cite{wu20153d} - ModelNet40 contains $12,311$ objects from $40$ categories, $9,843$ objects of which are for training and $2,468$ objects are for test. The object models are normalized to unit sphere, and $250,000$ query points and corresponding SDF values are sampled from the unit sphere for each object model, where the space near the mesh surface is sampled much denser than the space far away. $1,024$ points are sampled from the object model using Farthest Point Sampling (FPS) for object classification.

\vspace{0.2cm}
\noindent\textbf{ShapeNet} \cite{chang2015shapenet} - We only use the objects existing in ShapeNet Part \cite{Yi16} for training and testing. There are $16,880$ objects form $16$ categories in ShapeNet Part, $14,006$ objects of which are for training and $2,874$ objects are for test. The training data is obtained in the same way as ModelNet40. 
% For part segmentation, we use the point cloud data in \cite{Yi16}, where $50$ parts of the objects are annotated in total and there are no more than $6$ parts for each object.

\subsection{Implementation Details}
The input of ISSN and ISSN-CSL are $16,384$ points randomly sampled from $250,000$ query points of each object and the latent codes initialized with normal distribution. We train ISSN and ISSN-CSL with the batch size of $48$ on two 2080Ti GPU. The samples of category of interest and other categories are sampled equally in a mini-batch of ISSN-CSL. We optimize the weights in decoder and latent codes jointly using Adam optimizer with initial learning rate $5e^{-4}$ and $1e^{-3}$, respectively, for $500$ epochs (divided by $2$ every $200$ epochs). $\beta$ and $\delta$ are set to $0.1$. The values of $\lambda$ and $\gamma$ will be specified in the experiments. 
We will discuss how we set the values of hyper-parameters in Appendix \ref{sup:params}.

%For point cloud classification, we use the PyTorch implementations of DGCNN \cite{wang2019dynamic} and PointNet++ \cite{qi2017pointnet++} and keep the hyperparameters and optimizers consistent to ones in their the papers.

\subsection{Saliency Map Analysis}
% In this Section, we will present and analyze the saliency maps generated by different methods. Firstly, we will present quantitative comparisons of different saliency maps with respect to the metrics of saliency smoothness ratio and saliency symmetry distance.
In this Section, we will present quantitative comparisons of different saliency maps with respect to the metrics of saliency smoothness ratio and saliency symmetry distance.

\subsubsection{Smoothness Analysis}
% \
\quad We evaluate the smoothness of the saliency maps generated by different methods quantitatively using SSR. The results are shown in Table \ref{tab:modelnet_smooth} and Table \ref{tab:shapenet_smooth}. It clearly shows that the saliency maps of our proposed methods are smoother than those of PCSM \cite{zheng2019pointcloud} and PCA \cite{bae2008method}.

\begin{table}[htbp]
    \centering
        \begin{tabular}{  l | c | c | c  }
             % \hline
            \toprule[1pt]
            Methods         &  Mean   & Min    & Max   \\
            \hline
            PCA             &  0.119  &  0.024  &  0.352   \\
            PCSM            &  0.311  &  0.257  &  0.347   \\
            \hline
            ISSN ($\lambda=1e-3$)        &  0.089  &  0.012  &  0.315   \\
            ISSN-CSL ($\lambda=0,\gamma=1.0$)      &  0.059  &  0.011  &  0.392   \\
            ISSN-CSL ($\lambda=1e-3,\gamma=0.1$)        &  \textbf{0.042}  &  \textbf{0.009}  &  \textbf{0.178}   \\
            \bottomrule[1pt]
        \end{tabular}
    \caption{Statistical SSR of objects on ModelNet40 \cite{wu20153d}. Smaller is better for all columns.}
    \label{tab:modelnet_smooth}
    \vspace{-0.5em}
\end{table}

\begin{table}[htbp]
    \centering
        \begin{tabular}{  l | c | c | c  }
             % \hline
            \toprule[1pt]
            Methods         &  Mean   & Min     & Max    \\
            \hline
            PCA             &  0.083  &  0.025  &  0.214   \\
            PCSM            &  0.300  &  0.265  &  0.331   \\
            \hline
            ISSN ($\lambda=1e-3$)                 &  0.062  &  0.013  & 0.274   \\
            ISSN-CSL ($\lambda=0,\gamma=1.0$)      &  \textbf{0.033}  &  0.007  &  \textbf{0.228}   \\
            ISSN-CSL ($\lambda=1e-3,\gamma=0.1$)  &  0.043  &  \textbf{0.004}  &  0.243   \\
            \bottomrule[1pt]
        \end{tabular}
    \caption{Statistical SSR of objects on ShapeNet \cite{chang2015shapenet}. Smaller is better for all columns.}
    \label{tab:shapenet_smooth}
    \vspace{-0.5em}
\end{table}

\subsubsection{Symmetry Analysis}

\quad We evaluate the symmetry of the saliency maps on the symmetric objects quantitatively using our proposed symmetry distance. The results are shown in Table \ref{tab:modelnet_sym} and Table \ref{tab:shapenet_sym}. We can find that the saliency maps of our proposed methods can maintain better symmetry than those of PCSM \cite{zheng2019pointcloud} and PCA \cite{bae2008method}.

\begin{table}[htbp]
    \centering
        \begin{tabular}{  l | c | c | c  }
             % \hline
            \toprule[1pt]
            Methods         & Mean  & Min   & Max   \\
            \hline
            PCA             & 0.105 & 0.024 & 0.460  \\
            PCSM            & 0.305 & 0.234 & 0.346  \\
            \hline
            ISSN ($\lambda=1e-3$)                & \textbf{0.064} & 0.017 & \textbf{0.333}  \\
            ISSN-CSL ($\lambda=0,\gamma=1.0$)     & 0.161 & 0.026 & 0.492  \\
            ISSN-CSL ($\lambda=1e-3,\gamma=0.1$) & 0.157 & \textbf{0.016} & 0.507  \\
            \bottomrule[1pt]
        \end{tabular}
    \caption{Statistical symmetry distance of saliency map on ModelNet40 \cite{wu20153d}. Smaller is better for all columns.}
    \label{tab:modelnet_sym}
    \vspace{-0.5em}
\end{table}

\begin{table}[htbp]
    \centering
        \begin{tabular}{  l | c | c | c  }
             % \hline
            \toprule[1pt]
            Methods         & Mean  & Min   & Max   \\
            \hline
            PCA             & 0.144 & 0.007 & 0.408  \\
            PCSM            & 0.293 & 0.079 & 0.341  \\
            \hline
            ISSN ($\lambda=1e-3$)                & \textbf{0.067} & 0.016 & \textbf{0.295}  \\
            ISSN-CSL ($\lambda=0,\gamma=1.0$)     & 0.077 & 0.007 & 0.403  \\
            ISSN-CSL ($\lambda=1e-3,\gamma=0.1$) & 0.073 & \textbf{0.001} & 0.500  \\
            \bottomrule[1pt]
        \end{tabular}
    \caption{Statistical symmetry distance of saliency map on ShapeNet \cite{chang2015shapenet}. Smaller is better for all columns.}
    \label{tab:shapenet_sym}
    \vspace{-0.5em}
\end{table}

% \quad As shown in Figure \ref{fig:mesh_issn}, 

\begin{figure*}[htbp]
\centering
\subfloat[ISSN ($\lambda=0.001$)]{
\includegraphics[width=0.315\linewidth]{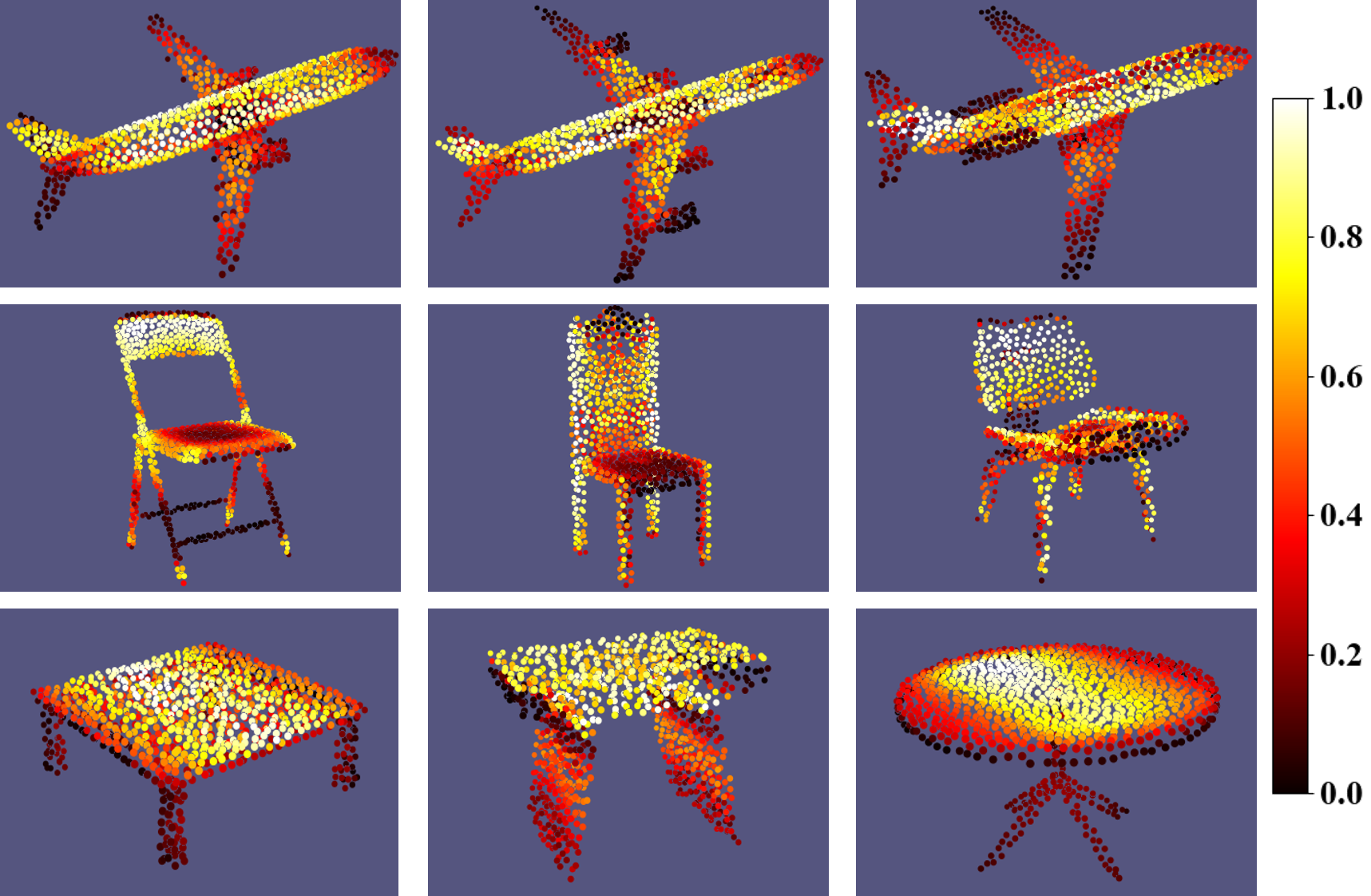} \label{fig:pc_issn}
% \caption{Vanilla}
}
\hspace{0.05cm}
\subfloat[ISSN-CSL ($\lambda=0,\gamma=1.0$)]{
\includegraphics[width=0.315\linewidth]{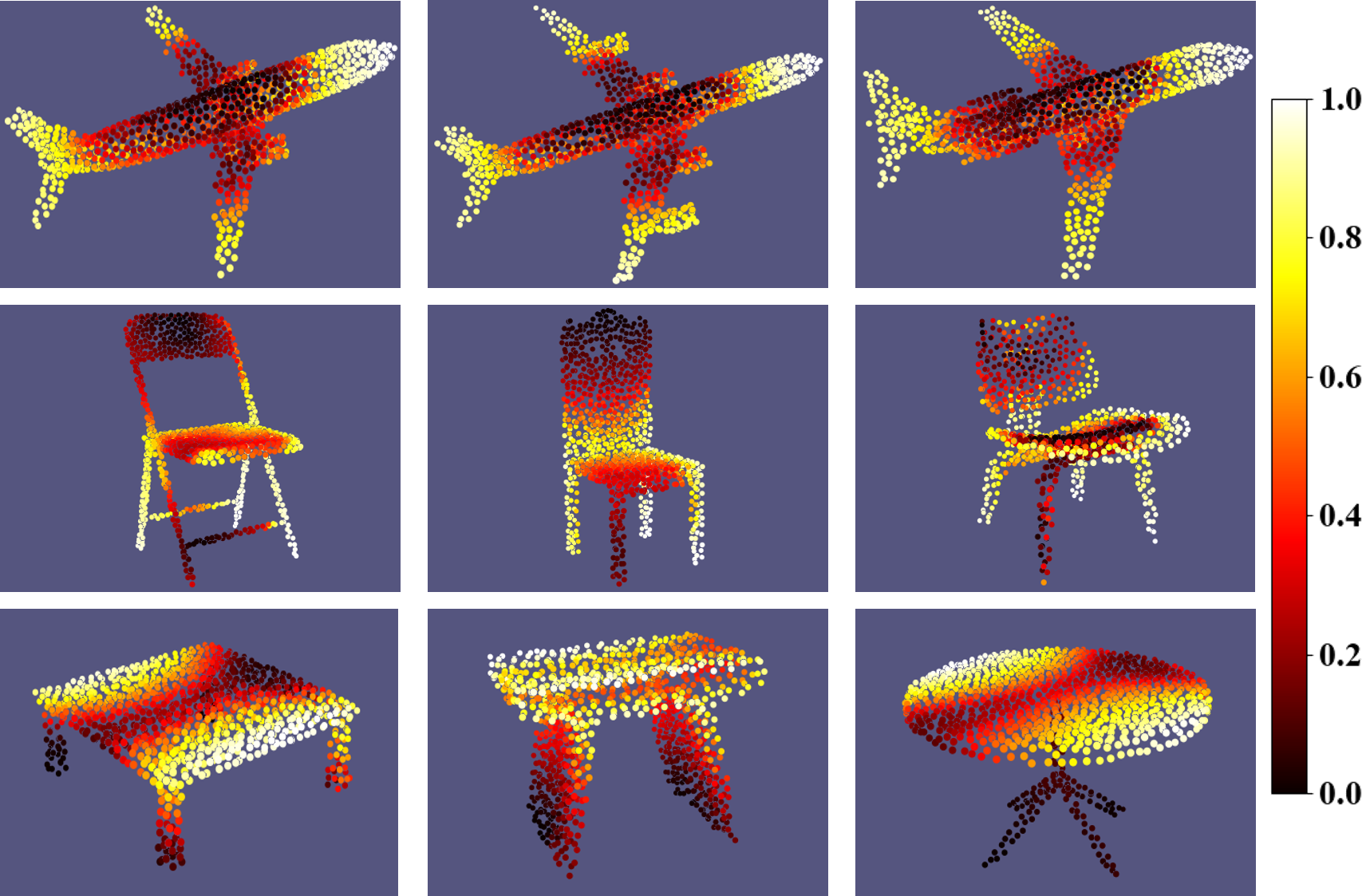} \label{fig:pc_csl}
% \caption{With contrastive classification network}
}
\hspace{0.05cm}
\subfloat[ISSN-CSL ($\lambda=0.001,\gamma=0.1$)]{
\includegraphics[width=0.315\linewidth]{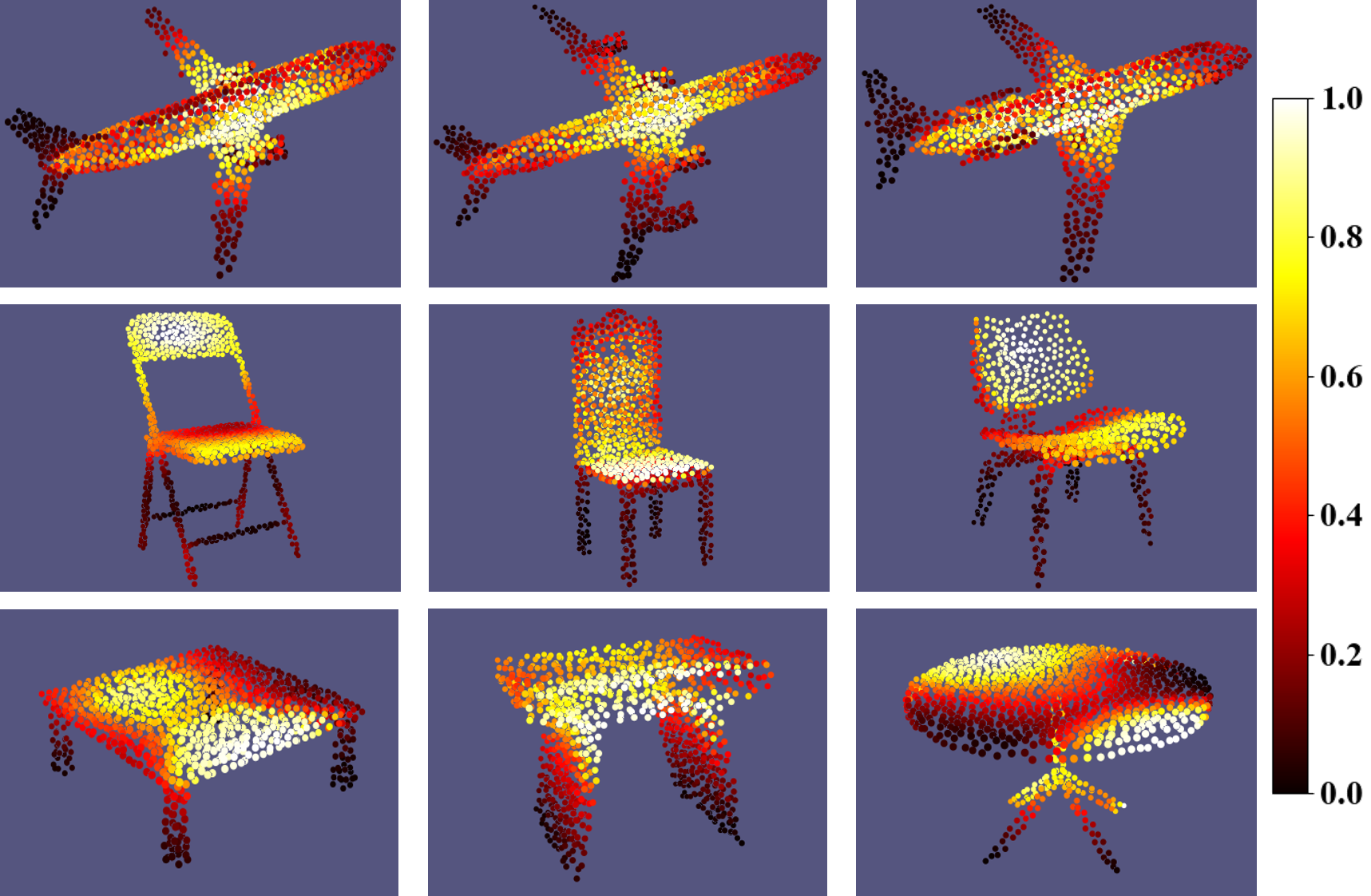} \label{fig:pc_com}
% \caption{With contrastive classification network}
}\\
% \hspace{0.03cm}
\subfloat[PCSM]{
\includegraphics[width=0.315\linewidth]{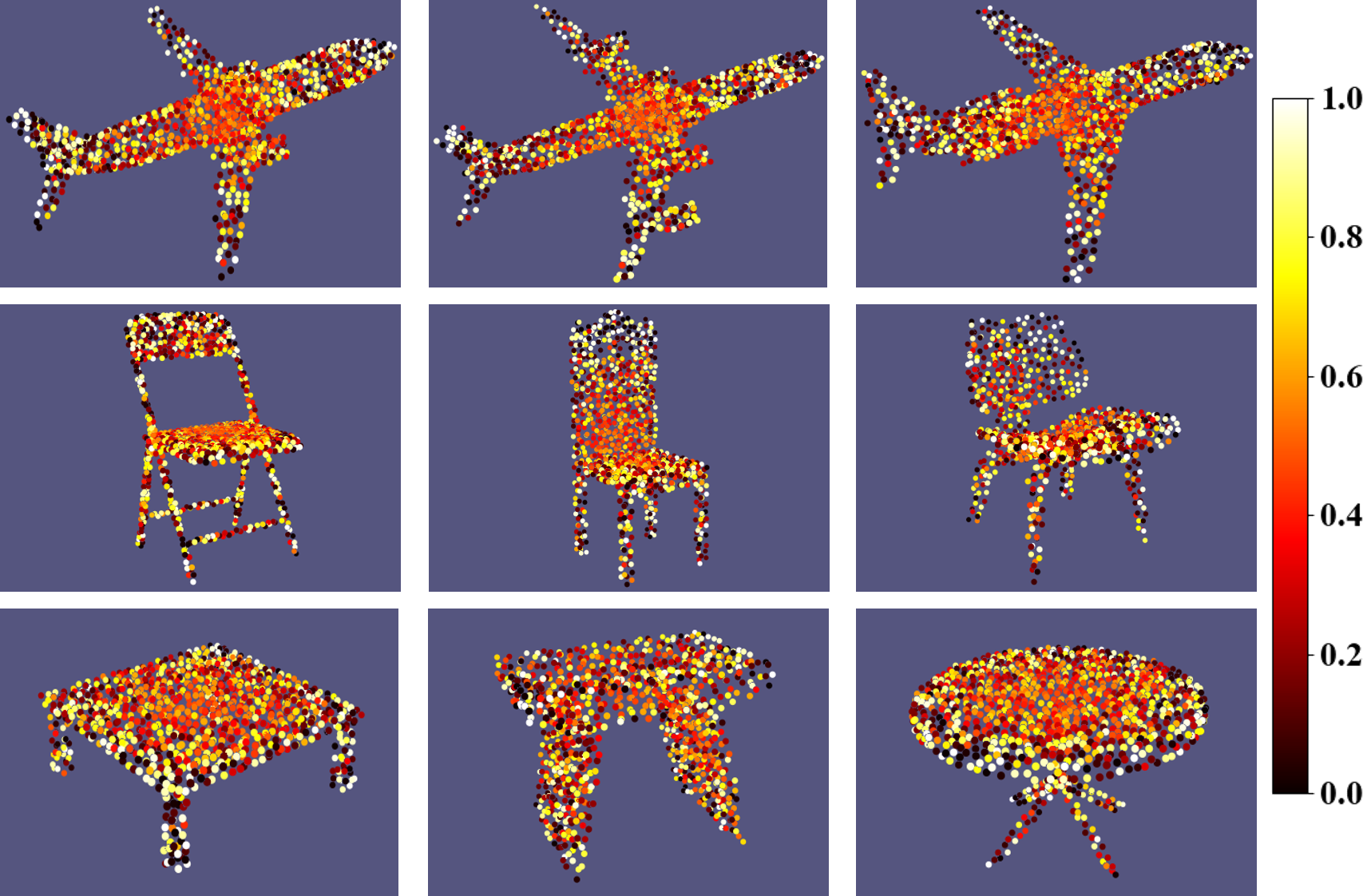}
% \caption{With contrastive classification network}
}
\hspace{0.05cm}
% \qquad\qquad
\subfloat[PCA]{
\includegraphics[width=0.315\linewidth]{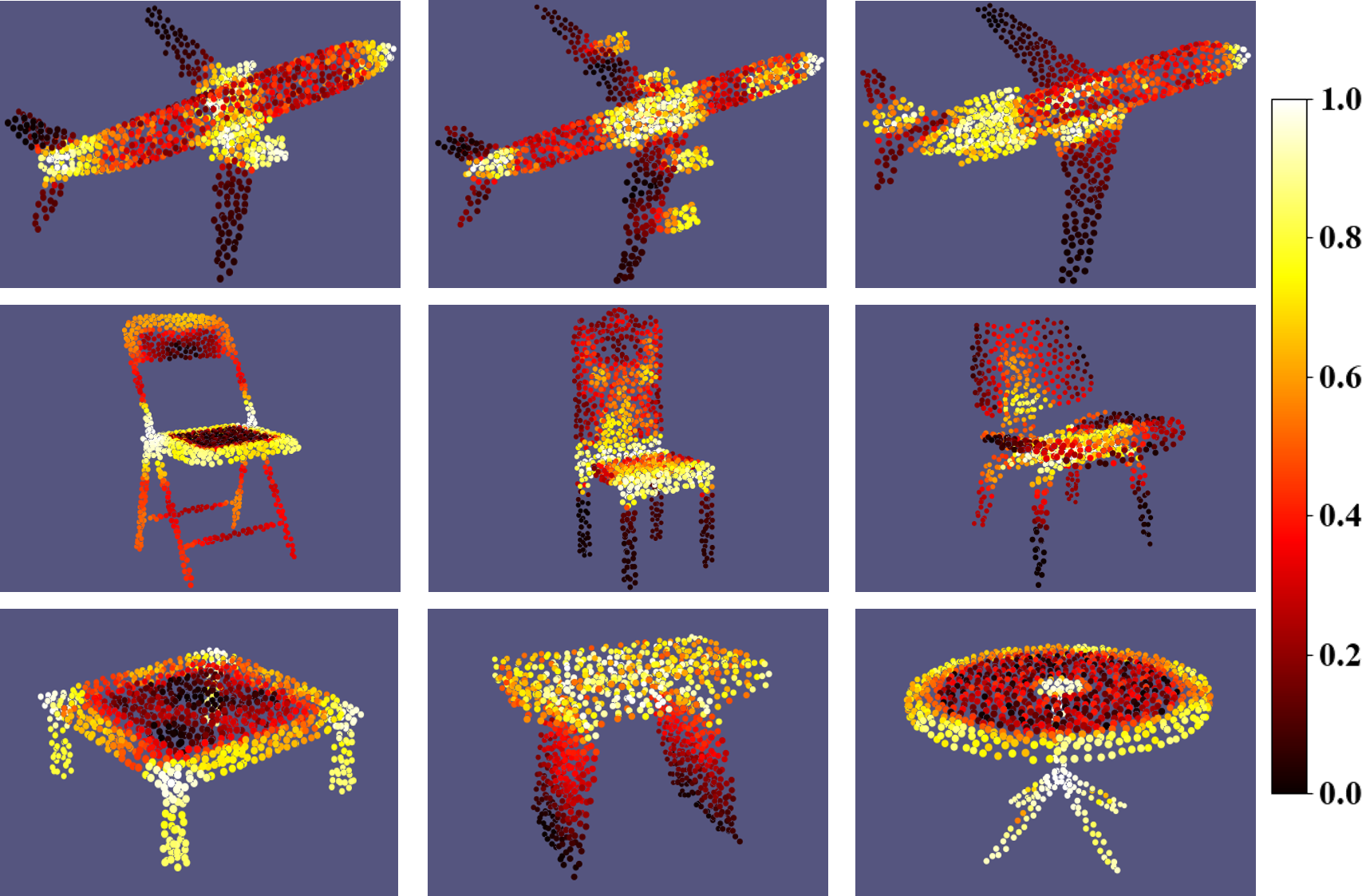}
% \caption{With contrastive classification network}
}
\hspace{1.0cm}
\subfloat[Symmetric object]{
\includegraphics[width=0.26\linewidth]{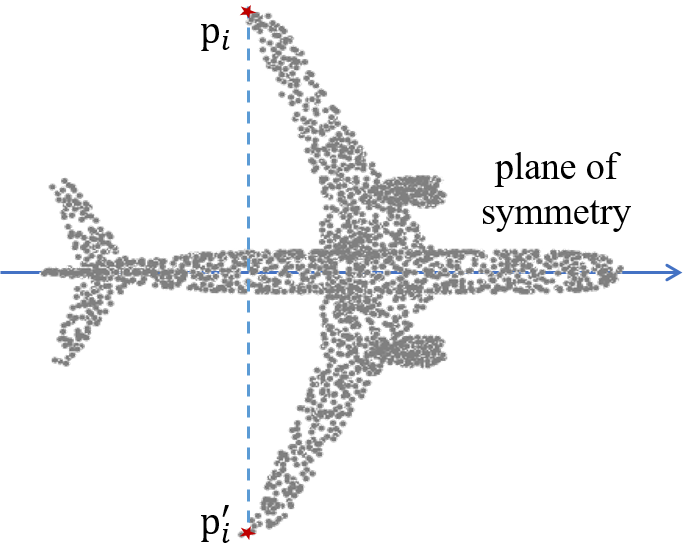} \label{fig:symmetry}
}
\caption{(a-e) Visualization of saliency maps of different methods on point cloud. These models are form ModelNet40 \cite{wu20153d} and the saliency maps of each object are normalized to $[0,1]$. (f) An example of symmetric object.}\label{fig:pc_sal}
\vspace{-0.8em}
\end{figure*}

\subsection{Visualization}
In this section, we will visualize the saliency maps on point clouds and analyze the saliency maps generated by different methods.

For the point cloud $\mathbf{P}$ sampled from the mesh surface, it's easy for us to obtain its saliency map by feeding the points in $\mathbf{P}$ and its corresponding latent code into the decoder directly. Figure \ref{fig:pc_sal} shows the saliency maps generated by our proposed ISSN and ISSN-CSL, PCSM \cite{zheng2019pointcloud}, and PCA \cite{bae2008method} on ModelNet40 \cite{wu20153d}. (More visualization results of our methods are shown in Appendix \ref{sup:visual}.) For fair comparisons, all of the saliency maps are normalized to $[0, 1]$. The figures demonstrate that the saliency maps of our proposed methods are much smoother than those of PCSM and PCA. Besides, the saliency maps of our proposed ISSN can maintain symmetry for those symmetric objects, which are even better than those of the analytic method, PCA. Although the saliency maps of our proposed ISSN-CSL are asymmetric on some categories and look different when $\lambda$ and $\gamma$ are assigned with different values, but they can capture consistent pattern across different objects from the same category. Contrarily, saliency maps generated by PCSM look noisy and asymmetric and we can not find any regular pattern among them.

Apart from the properties of smoothness and symmetry, the saliency maps generated by our proposed ISSN indeed capture the representativeness and common parts shared among the objects of the same category. For example, all of the airplanes in Figure \ref{fig:pc_issn} have similar fuselages and wings but different engines and empennages. As a result, the saliency scores on the fuselages and wings are much higher than those on the engines and empennages. The regular pattern can also be found on the objects from other category, \eg the chairs and the tables in Figure \ref{fig:pc_issn}. Moreover, the saliency map generated by ISSN-CSL can capture discriminations among different categories. As shown in \ref{fig:pc_csl}, both the chairs and tables have legs, to avoid confusing, it guides the saliency maps of chairs to focus on their legs and the saliency maps of tables to focus on their tops. The saliency map of both ISSN and ISSN-CSL can capture consistent pattern across different objects from the same category. In Section \ref{exp_cls}, we will show their potential on improving the performance of point cloud classification.

\section{Applications}

% A smooth and symmetry saliency map will provide great contributions on shape simplification and point cloud classification. 
A smooth and symmetry saliency map with semantic representations will contribute to the reconstruction of category-salient and instance-specific parts and point cloud classification.

\begin{figure*}[htbp]
\centering
\subfloat[ISSN ($\lambda=0.001$)]{
\includegraphics[width=0.315\linewidth]{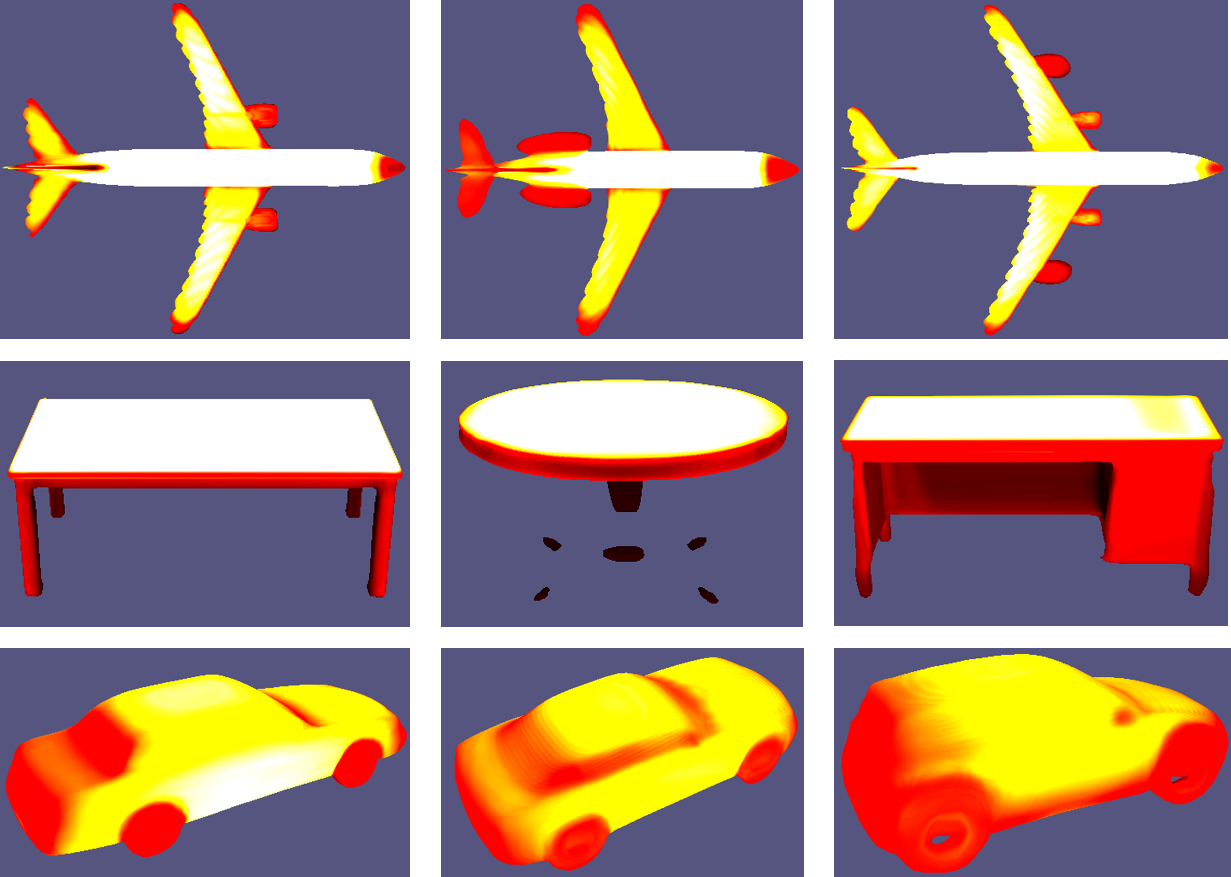} \label{fig:mesh_issn}
% \caption{Vanilla}
}
\hspace{0.03cm}
\subfloat[ISSN-CSL ($\lambda=0,\gamma=1.0$)]{
\includegraphics[width=0.315\linewidth]{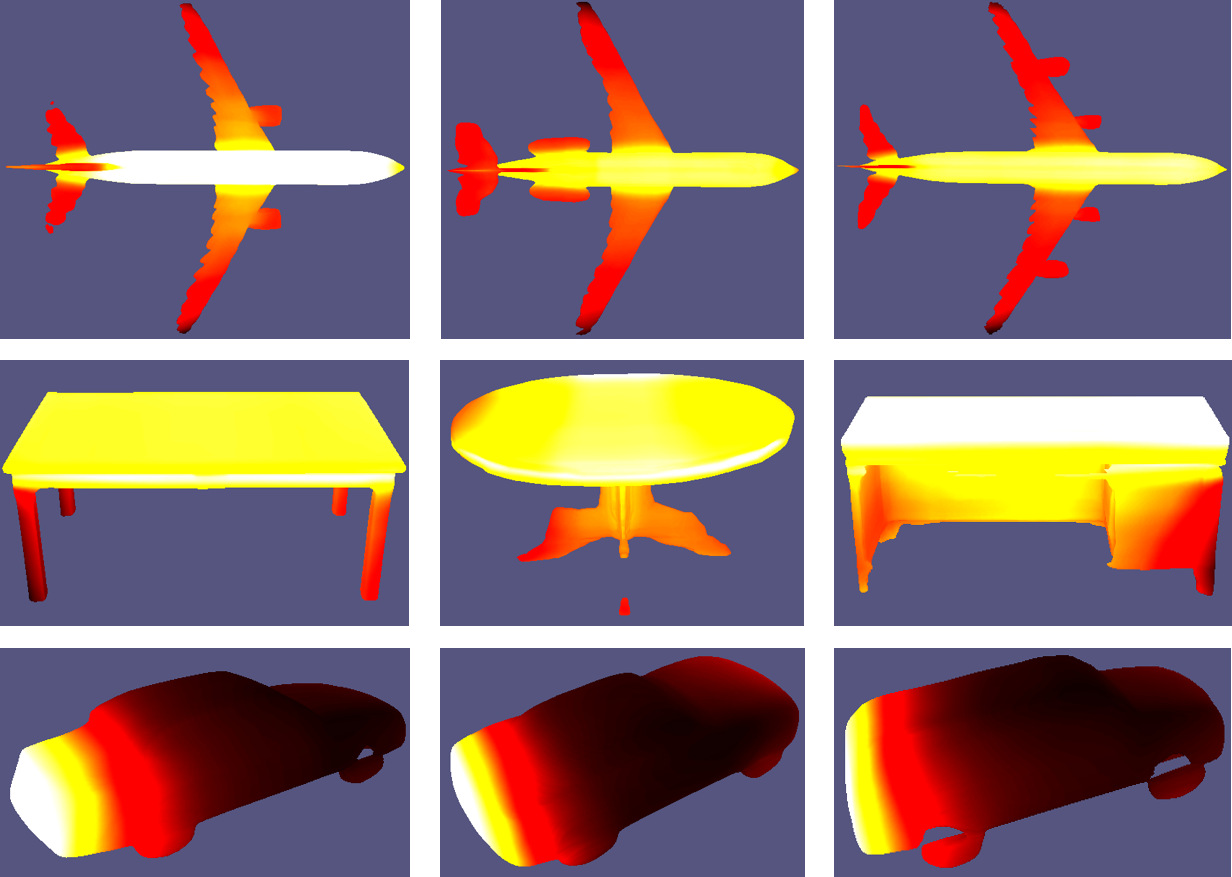} \label{fig:mesh_csl}
% \caption{With contrastive classification network}
}
\hspace{0.03cm}
\subfloat[ISSN-CSL ($\lambda=0.001,\gamma=0.1$)]{
\includegraphics[width=0.315\linewidth]{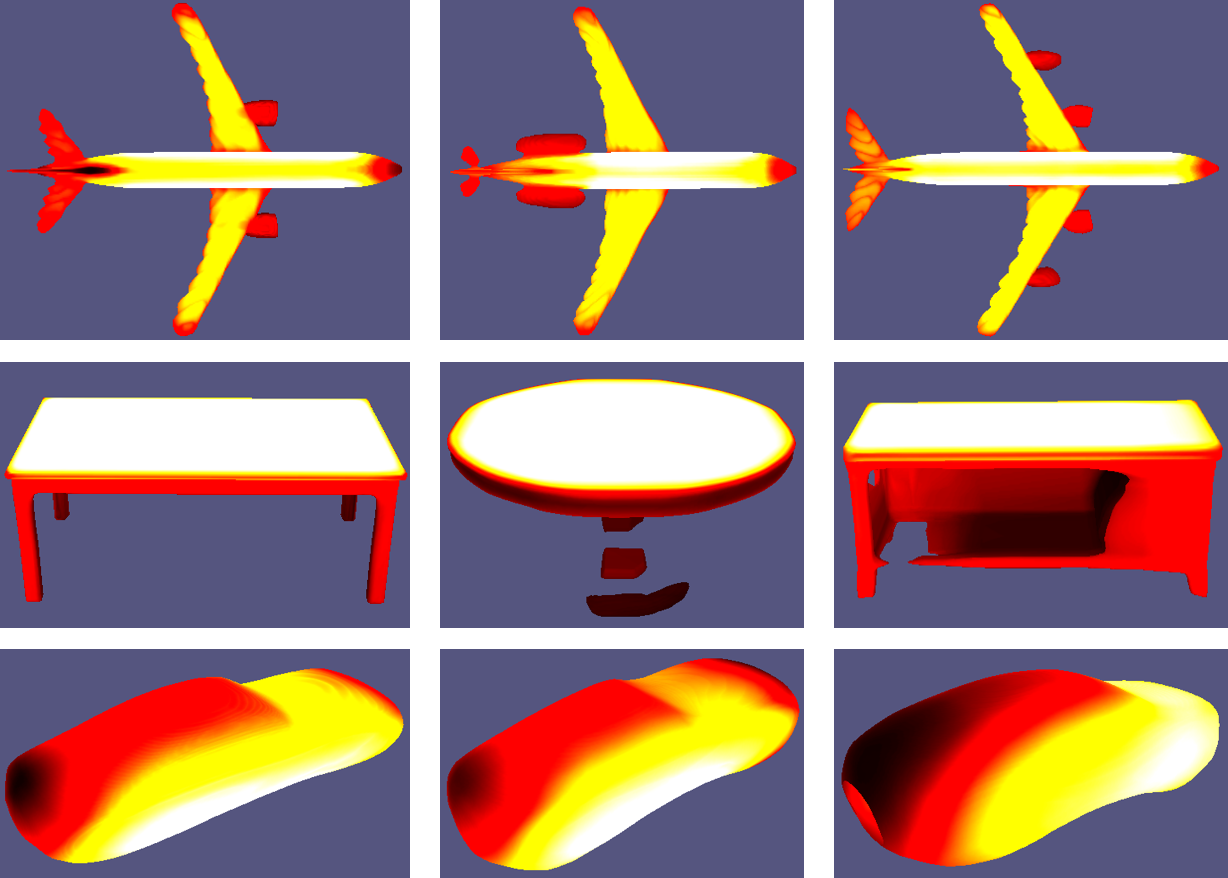} \label{fig:mesh_com}
% \caption{With contrastive classification network}
}
\caption{Visualization of saliency maps on reconstructed meshes by our proposed methods. The brighter color denotes higher saliency. These models are from ShapeNet \cite{chang2015shapenet}.}\label{fig:mesh_sal}
\vspace{-0.8em}
\end{figure*}

\subsection{Reconstruction}
During test phase, we sample $100,000$ query points near the surfaces of objects and calculate corresponding SDF values with the help of point clouds and normals provided in \cite{Yi16}. Then, we fix the weights of decoder in DeepSDF and use the sampled query points and SDF values to optimize the latent codes.
With the trained decoder and optimized latent codes, we feed the coordinates of a normalized 3D grid with resolution of $256^3$ into the network to obtain the SDF values and corresponding saliency scores. 
Then, we apply the Marching Cubes algorithm to reconstruct the mesh model. Note that the saliency scores are interpolated together with the SDF values when searching the vertices on the zero iso-surface in the Marching Cubes algorithm. The reconstructed meshes and saliency maps on ShapeNet \cite{chang2015shapenet} are shown in Figure \ref{fig:mesh_sal}. We can find that ISSN and ISSN-CSL can learn to generate continuous and smooth saliency map along the manifold of the object surface and maintain the symmetry property for those symmetric objects. 

\begin{figure}[htbp]
\centering
\subfloat[Airplane]{
\includegraphics[width=0.9\linewidth]{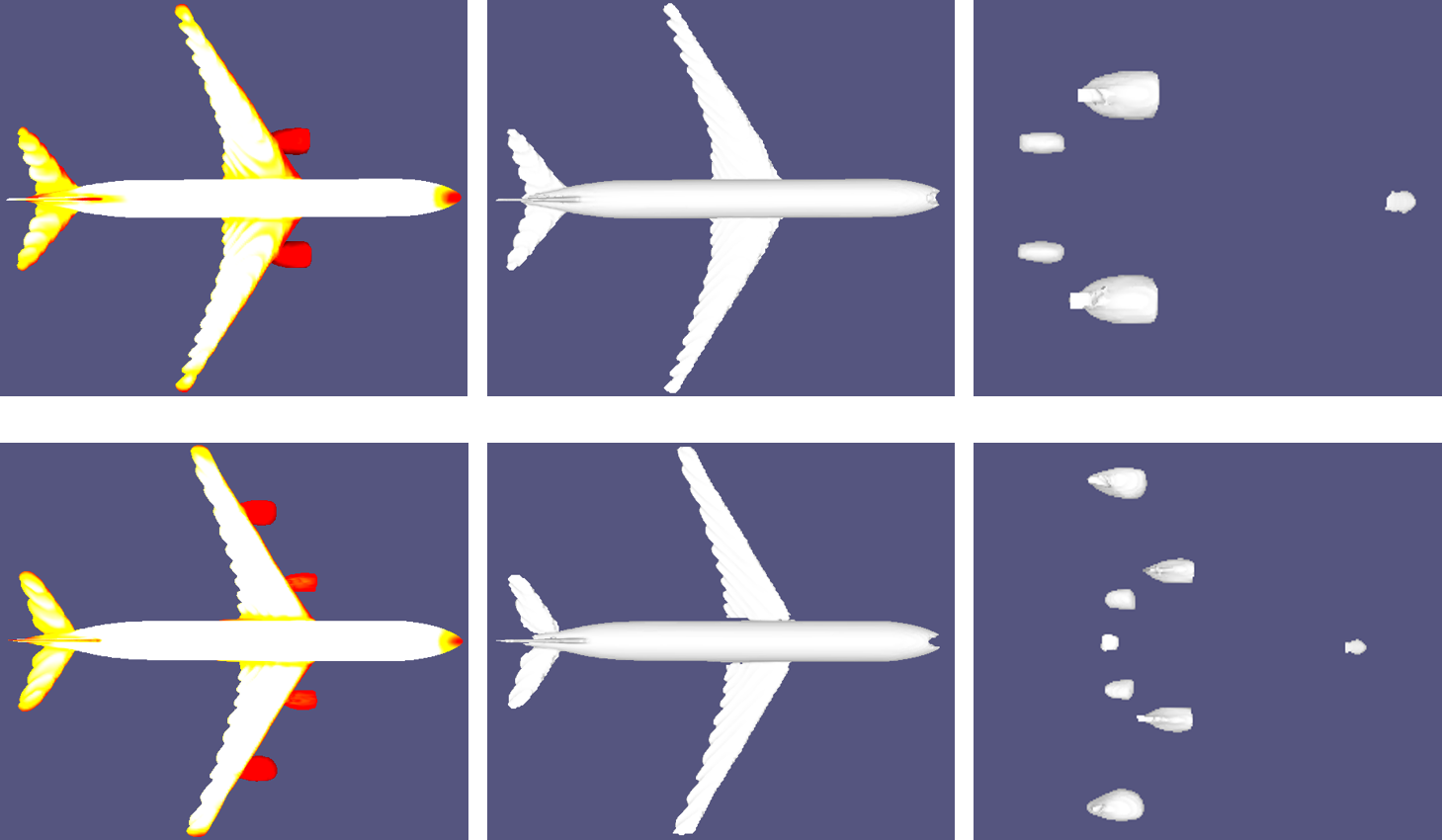}
% \caption{Vanilla}
}\\
% \hspace{0.1cm}
\subfloat[Mug]{
\includegraphics[width=0.9\linewidth]{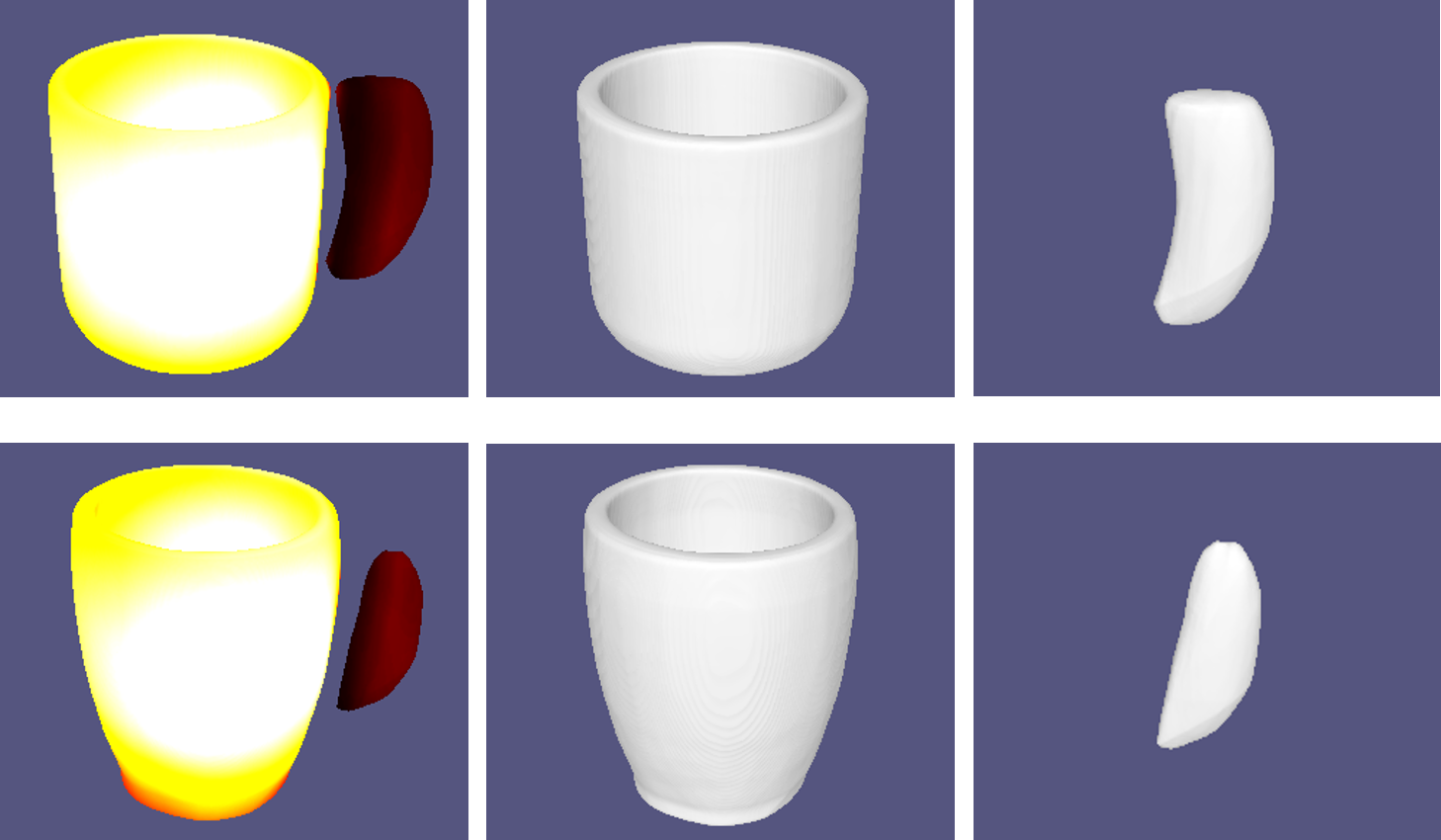}
% \caption{With contrastive classification network}
}\\
% \hspace{0.1cm}
\subfloat[Skateboard]{
\includegraphics[width=0.9\linewidth]{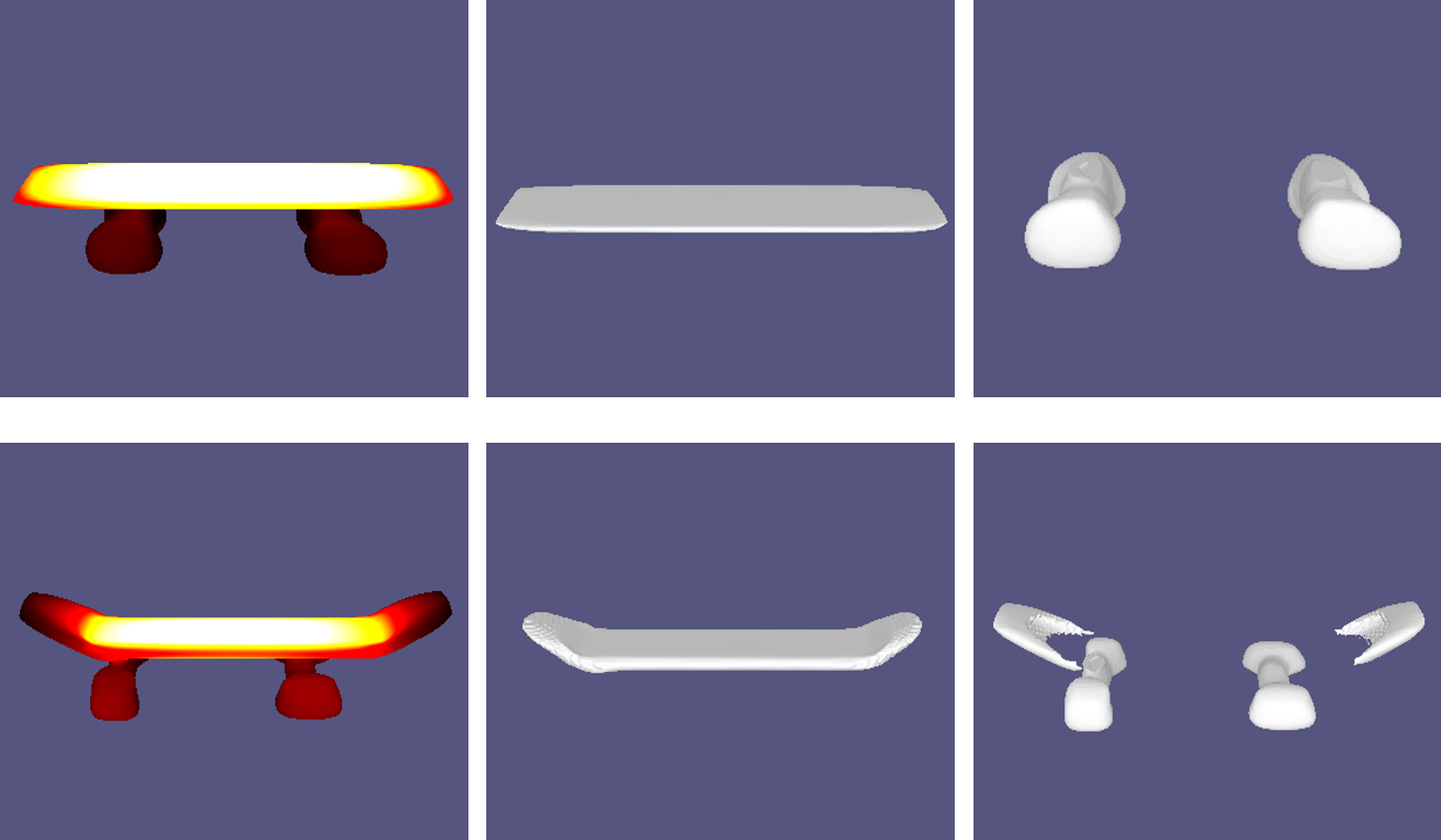}
% \caption{With contrastive classification network}
}
\caption{Reconstruction results. Saliency maps (left), category-salient parts (middle), and instance-specific parts (right).}\label{fig:recon}
\vspace{-0.8em}
\end{figure}

Another straightforward application of the saliency map of our proposed ISSN is to reconstruct the category-salient parts and the instance-specific parts, respectively. We remove those points whose saliency scores are below a threshold, then apply the Marching Cubes algorithm to reconstruct the common parts shared among the objects of the same category. We can also obtain the instance-specific parts by removing those points whose saliency scores are above a threshold. The results are shown in Figure \ref{fig:recon}.

\subsection{Point Cloud Classification} \label{exp_cls}
To verify the efficacy of saliency maps generated by our proposed ISSN-CSL on point cloud classification, we use the saliency maps on training set of ModelNet \cite{wu20153d} to modify the backward gradient of PointNet++ \cite{qi2017pointnet++} and DGCNN \cite{wang2019dynamic}. We firstly train the saliency networks for each category on training set, and use the trained networks to predict the saliency map for each sampled point cloud on training set. Since that we do not know the label of point cloud on test set, we introduce the following strategy to use the saliency map on training set to train the classification network. 

We denote by $\mathbf{P}=\{\mathbf{p}_i\}_{i=1}^n$ the object point cloud, $\mathbf{S}=\{s_i\}_{i=1}^n$ its saliency map, $\mathbf{f}_{mi}$ the feature of $\mathbf{p}_i$ at layer $m$ and $\mathcal{L}_{cls}$ the classification loss. Then we replace the derivative of $\mathcal{L}_{cls}$ with respect to $\mathbf{f}_{mi}$ with $\mathbf{g}_{mi}=s_i\frac{\partial \mathcal{L}_{cls}}{\partial \mathbf{f}_{mi}}$ and use $\mathbf{g}_{mi}$ to update the learnable weights at layer $m$. We use the PyTorch implementations of DGCNN \cite{wang2019dynamic} and PointNet++ \cite{qi2017pointnet++} and keep the hyperparameters and optimizers consistent to ones in their papers.

The results are shown in Table \ref{tab:cls}, we can see that the performances on both PointNet++ and DGCNN can be improved consistently when trained with saliency maps.

\begin{table}[htbp]
    \centering
        \begin{tabular}{  l | c | c }
             % \hline
            \toprule[1pt]
            \multirow{2}{*}{Methods} & \multicolumn{2}{c}{Accuracy ($\%$)} \\
            \cline{2-3}
                                      & Overall & Avg. class \\
            \hline
            PointNet++~\cite{qi2017pointnet++}    & 90.7  & - \\
            PointNet++ (our Exp.)                  & 92.5  & 90.4 \\
            PointNet++ w/ saliency                   & \textbf{93.3}  & \textbf{91.3} \\
            % PointNet2+sal ($\beta=0.1$)          & 86.1 & 84.5 \\
            % PointNet2+sal ($\beta=1.0$)          & 86.1 & 84.7 \\
            % \hline
            \hline
            DGCNN~\cite{wang2019dynamic} & 92.9 & 90.2 \\
            DGCNN (our Exp.)              & 93.3 & \textbf{91.0} \\
            DGCNN w/ saliency               & \textbf{93.7} & 90.4 \\
            \bottomrule[1pt]
        \end{tabular}
    \caption{Classification results on ModelNet40~\cite{wu20153d}.}
    \label{tab:cls}
    \vspace{-0.8em}
\end{table}

% \subsubsection{Saliency Points Clustering}

\begin{figure}[htbp]
\centering
\subfloat[ISSN]{
\includegraphics[width=0.49\linewidth]{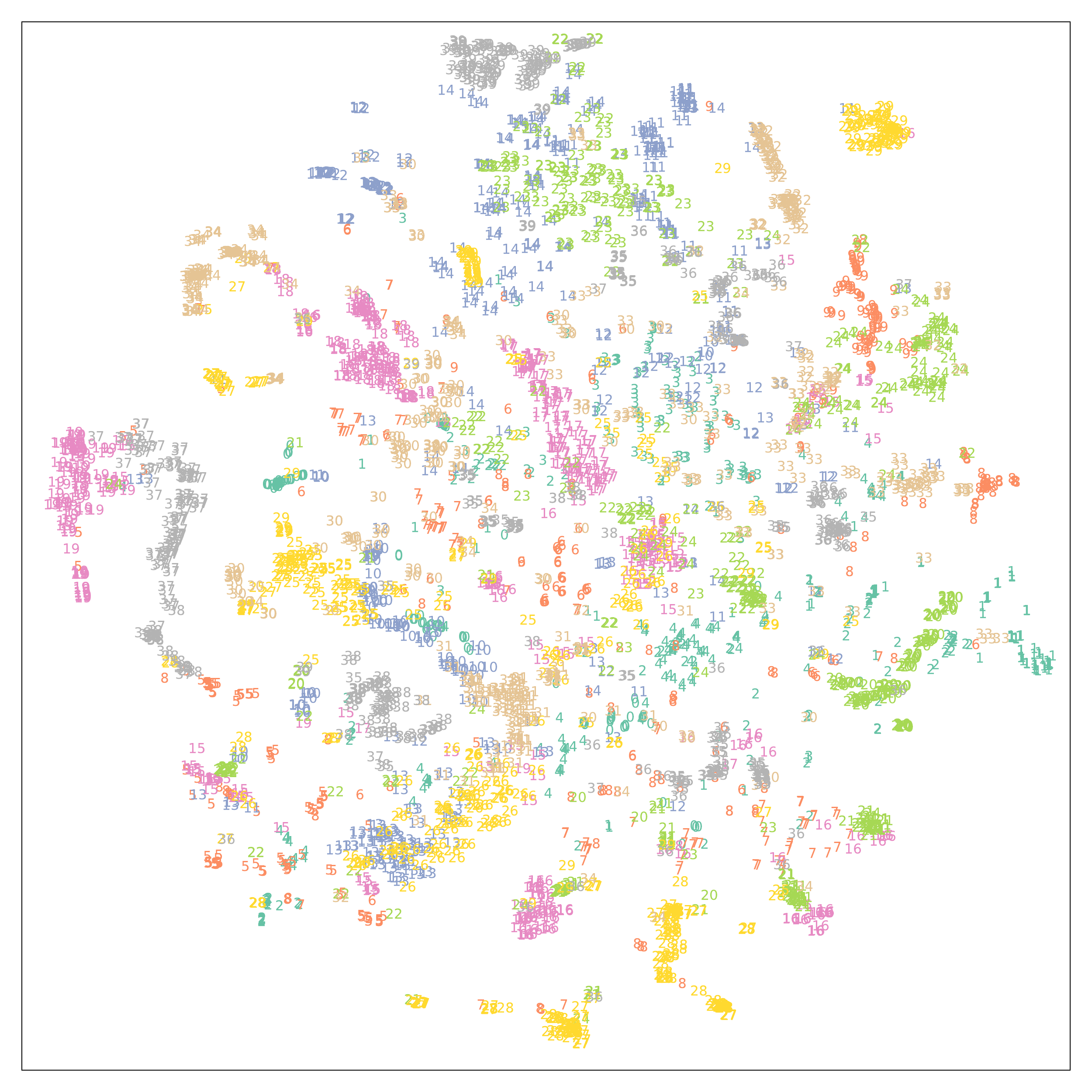}
% \caption{Vanilla}
}
\subfloat[ISSN-CSL]{
\includegraphics[width=0.49\linewidth]{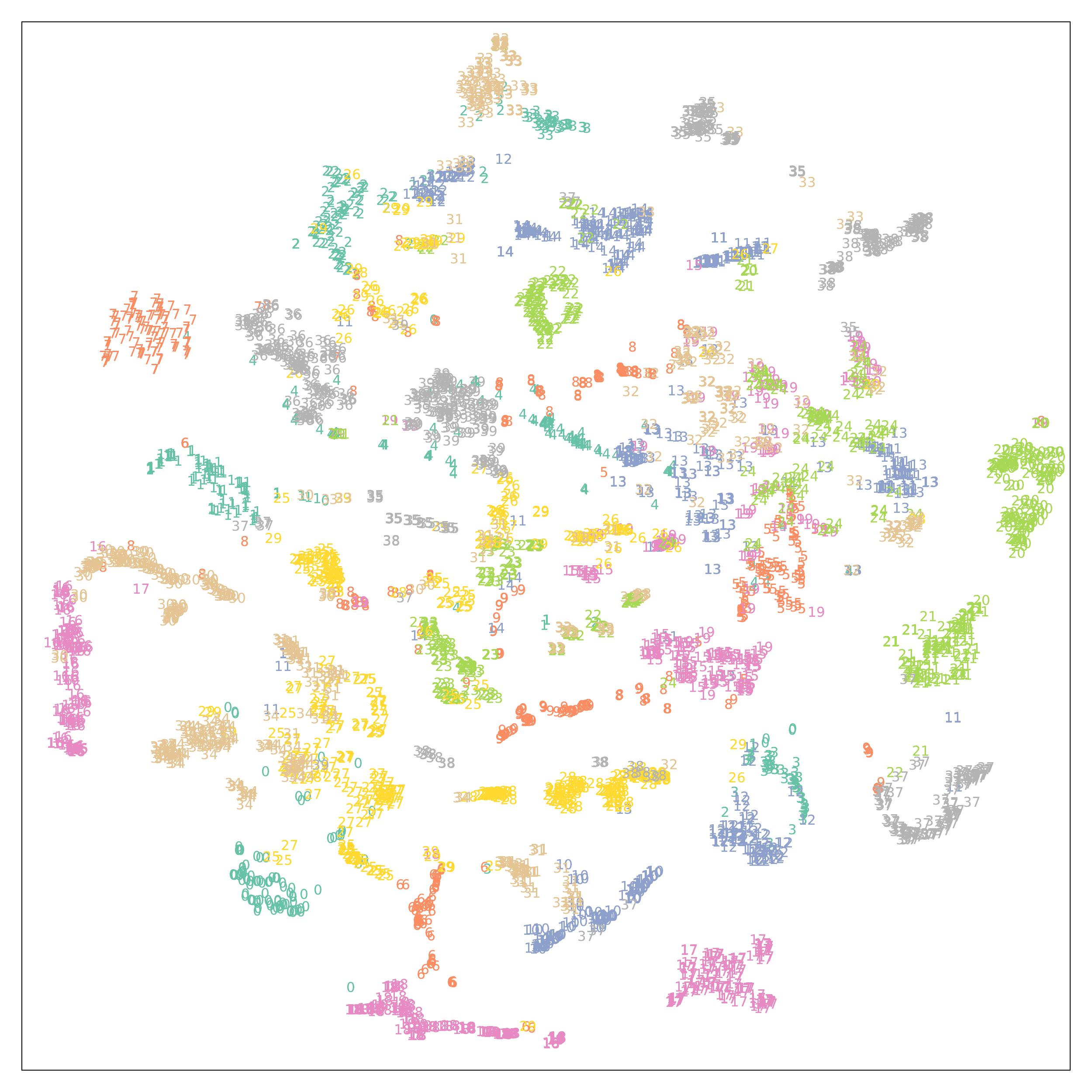}
% \caption{Vanilla}
}\\
% \quad
\subfloat[PCSM]{
\includegraphics[width=0.49\linewidth]{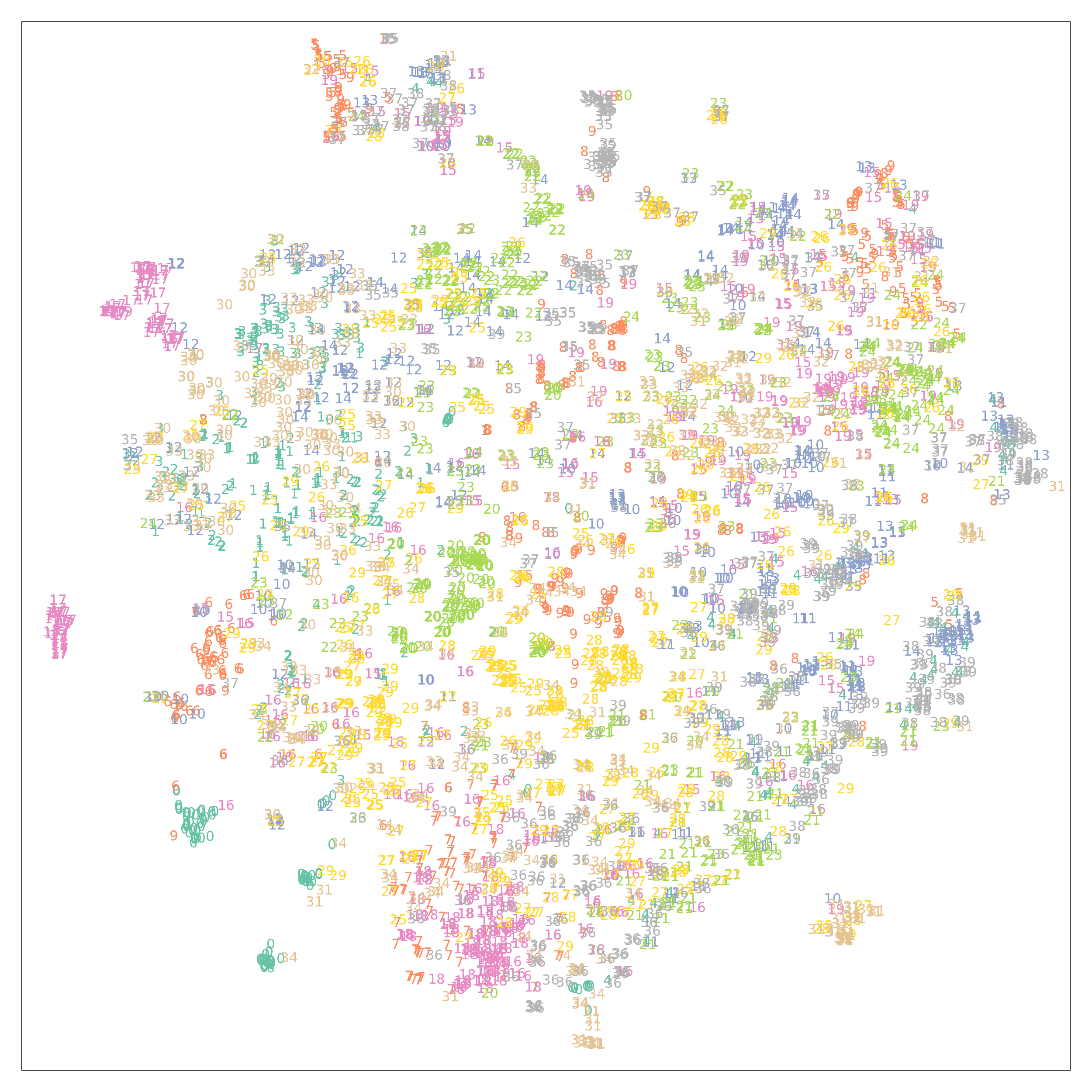}
% \caption{With contrastive classification network}
}
% \quad
\subfloat[PCA]{
\includegraphics[width=0.49\linewidth]{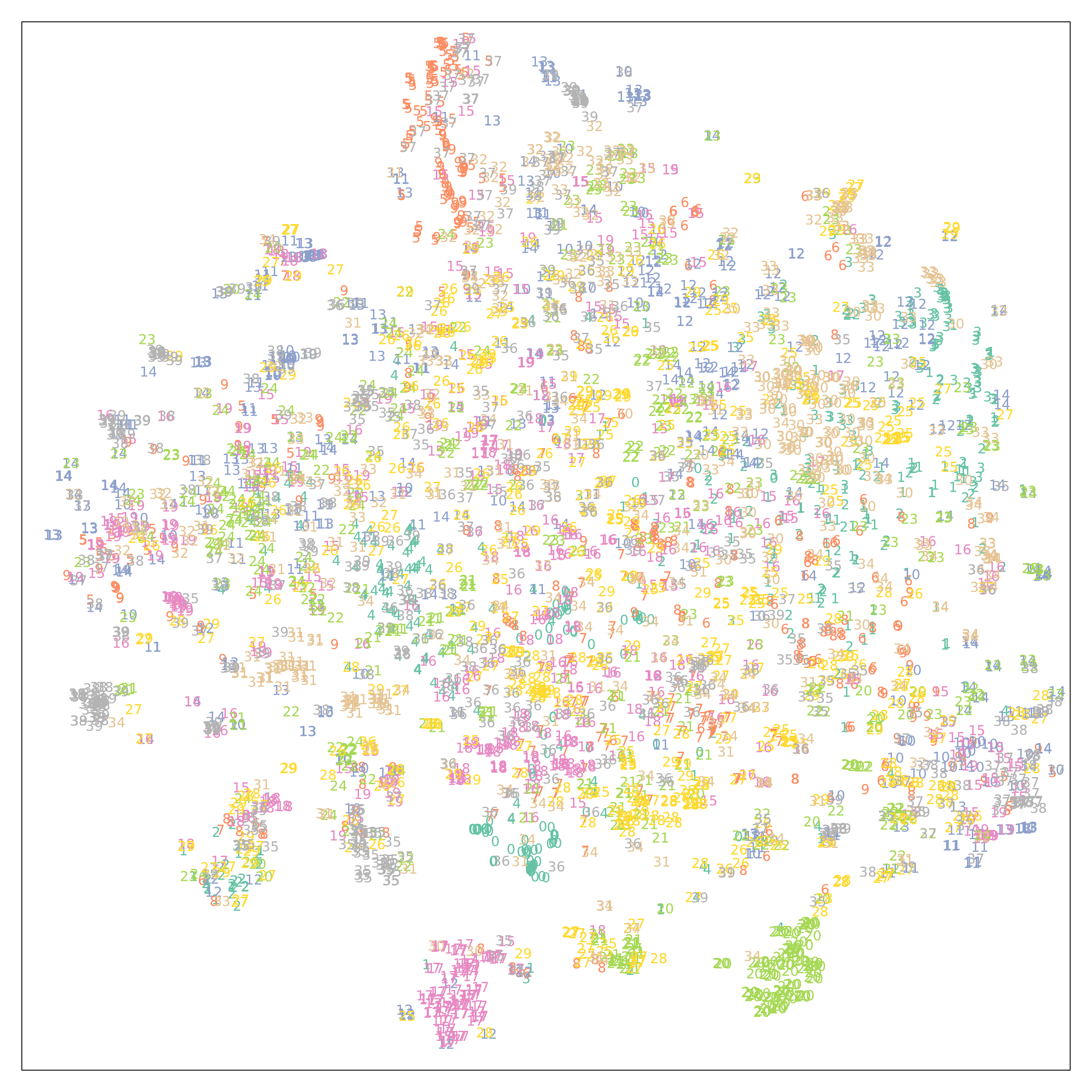}
}
\caption{t-SNE clustering results of saliency points (zoom in for details, the numbers in the figures correspond to categories in ModelNet40).}\label{fig:cluster}
\vspace{-0.5em}
\end{figure}

In order to show the representativeness and discrimination of the saliency maps estimated by different methods, we conduct the saliency points clustering experiment. Specifically, for the point cloud of each object in ModelNet40 \cite{wu20153d}, we obtain its saliency maps using our proposed ISSN and ISSN-CSL, PCSM \cite{zheng2019pointcloud}, and PCA\cite{bae2008method}. Then, we rank the points according to the saliency scores and select the top-$10$ points as the saliency points of the object. We utilize t-SNE \cite{maaten2008visualizing} to clutter and visualize the saliency points of different objects. Note that we use the Hausdorff distance to calculate the distance between two sets of saliency points. The clustering results are illustrated in Figure \ref{fig:cluster}. Our proposed ISSN-CSL presents the best clustering results, concentrating within the same category and discriminating between different category. Surprisingly, although we do not guide the ISSN to learn discrimination of instances from different categories, the clustering result is better than that of PCSM where the saliency maps are estimated via semantic classification networks.

Additionally, to compare the representativeness and discrimination of the saliency map estimated by different methods quantitatively, we design the saliency points classification task. In particular, we select the top-$k$ salient points of each object in both training set and test set of ModelNet40 \cite{wu20153d}. Then, we use the salient points in training set to train DGCNN \cite{wang2019dynamic} and evaluate the overall accuracy on test set. The results are shown in Table \ref{tab:sal_cls}. We can find that our proposed ISSN and ISSN-CSL outperform PCSM and PCA with a large margin, which are consistent to the clustering results. We also find that after dropping the redundant points based on the saliency maps of ISSN and ISSN-CSL, the performances of point classification can be improved a lot compared to the baseline using $1,024$ points (overall accuracy is $93.3\%$ in our experiment). The performances of PCSM and PCA on saliency points classification degrade dramatically when the number of saliency points decreases, while the performances of ISSN and ISSN-CSL are keeping robust.
The training and testing curves are shown in Appendix \ref{sup:curves}.

\begin{table}[htbp]
    \centering
        \begin{tabular}{  l | c | c | c | c | c }
             % \hline
            \toprule[1pt]
            \# Saliency points &   32   &  64   &  128   &  256   &   512    \\
            \hline
            PCA               &  75.3  &  79.5 &  84.1  &  87.0  &   91.1   \\
            PCSM               &  85.6  &  87.9 &  90.7  &  92.3  &   92.9   \\
            \hline
            ISSN           &  96.5  &  98.0 &  98.2  &  98.6  &   99.0   \\
            ISSN-CSL           &  \textbf{98.0}  &  \textbf{98.8} &  \textbf{99.4}  &  \textbf{99.6}  &   \textbf{99.7}   \\
            \bottomrule[1pt]
        \end{tabular}
    \caption{Overall accuracy ($\%$) of saliency points classification on ModelNet40~\cite{wu20153d} using DGCNN \cite{wang2019dynamic}.}
    \label{tab:sal_cls}
    \vspace{-1em}
\end{table}

\section{Conclusion}
% This paper exploits and explores a fundamental problem for 3D object shapes. This is the first attempt in this direction. Given the large shape variations among different instances of a same category, our proposed ISSN and ISSN-CSL quantity specifies how individual surface points contribute to the formation of the shape as the category or instance-specific parts of object surfaces. Our proposed methods have the properties of smoothness, symmetry, and semantic representativeness which contribute to the better shape reconstruction and point cloud classification. 
This paper exploits and explores a fundamental problem about what defines a category of object shapes. This is the first attempt in this direction. Given the large shape variations among different instances of a same category, our proposed ISSN and ISSN-CSL specify how individual surface points contribute to the formation of the shape as the category and the distinction between these instances and those of other categories, respectively. The learned saliency maps of our proposed methods have the properties of smoothness, symmetry, and semantic representativeness which contribute to reconstruction of either category-salient or instance-specific parts of object surfaces and point cloud classification. 
% \newpage
{\small
\bibliographystyle{ieee_fullname}
\bibliography{saliency}
}

\newpage

% \newpage

\appendix
% \clearpage
% \begin{center}
\noindent\textbf{\Large Appendix}
% \end{center}
\vspace{0.5cm}

%------------------------------------------------------------------------
In this appendix, we will provide more experiments to explain the hyper-parameters selection, present more saliency maps visualizations and training and testing curves on the saliency point classification. 

\section{Hyper-parameter Settings}\label{sup:params}

\subsection{$\lambda$ in ISSN}\label{sup:issn}
\begin{figure}[htbp]
\centering
\subfloat[$\lambda>=1e-2$]{
\includegraphics[width=0.90\linewidth]{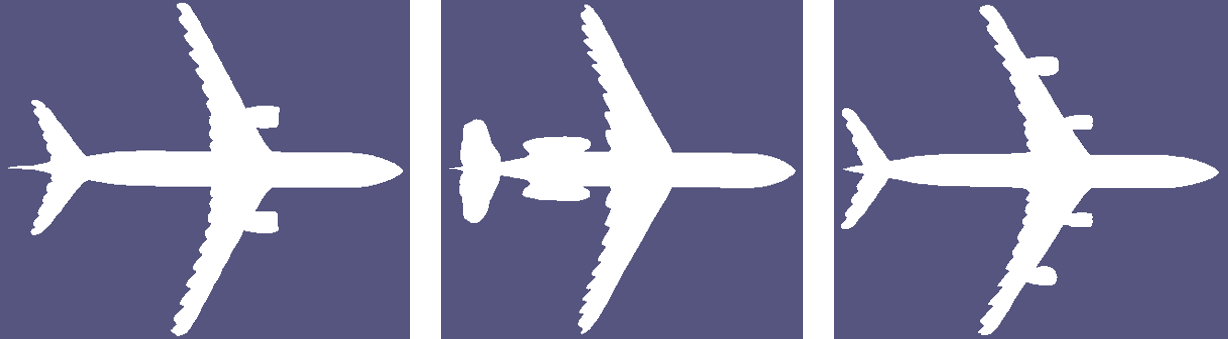}
}\\
\subfloat[$\lambda=1e-3$]{
\includegraphics[width=0.90\linewidth]{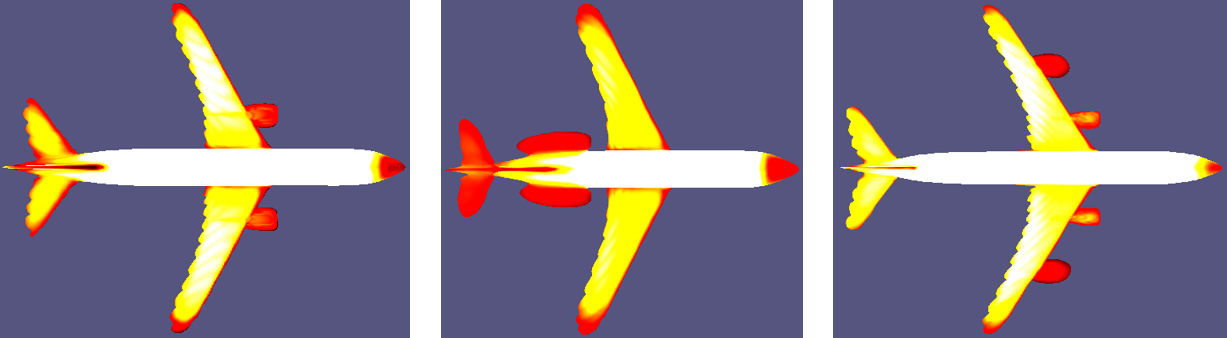}
}\\
\subfloat[$\lambda=1e-4$]{
\includegraphics[width=0.90\linewidth]{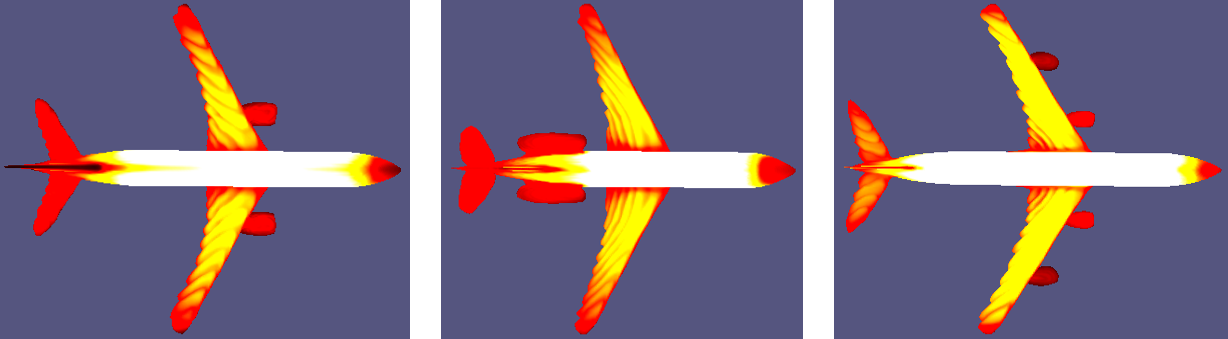}
}
\caption{The influence of $\lambda$ in ISSN.}\label{fig:lambda}
\vspace{-0.5em}
\end{figure}

We try different values of $\lambda$ in ISSN, the results are shown in Figure \ref{fig:lambda}. We find that when $\lambda>=1e-2$, the learned saliency maps all have high values over the object surfaces and can not capture differences between the category-salient parts and the instance-specific parts. When $\lambda=1e-4$, the learned saliency maps will become noisy. When $\lambda<1e-4$, the saliency scores will be vanished and even affect the SDF learning. Therefore, in the paper we set $\lambda=1e-3$ to learn better saliency representations.

\subsection{$\lambda$ and $\gamma$ in ISSN-CSL}

\begin{figure}[htbp]
\centering
\subfloat[$\gamma=1.0$]{
\includegraphics[width=0.90\linewidth]{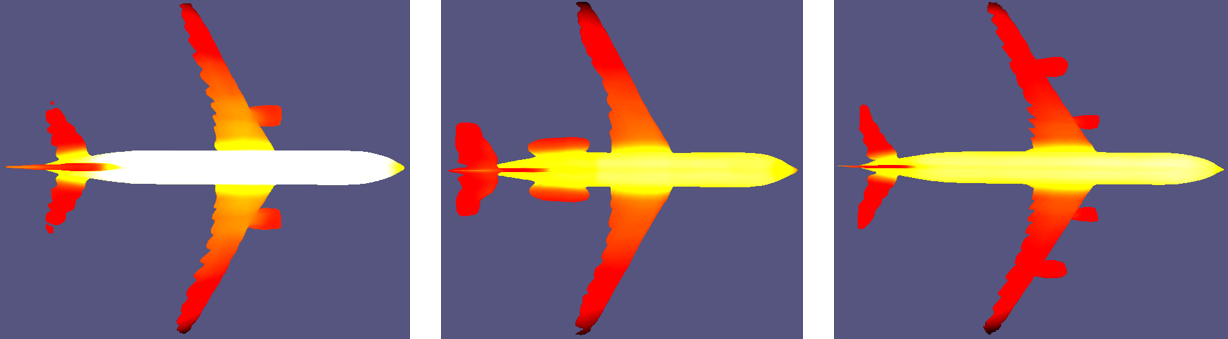}
}\\
\subfloat[$\gamma=1e-1$]{
\includegraphics[width=0.90\linewidth]{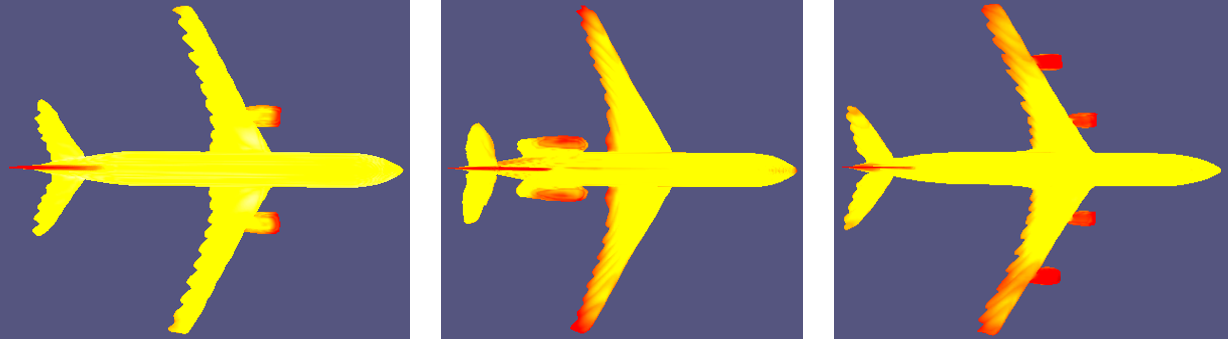}
}\\
\subfloat[$\gamma=1e-2$]{
\includegraphics[width=0.90\linewidth]{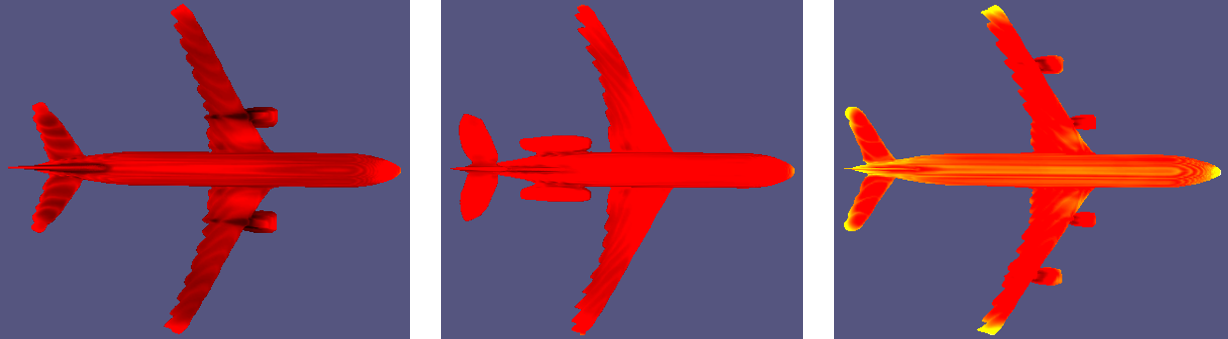}
}\\
\subfloat[$\gamma=1e-3$]{
\includegraphics[width=0.90\linewidth]{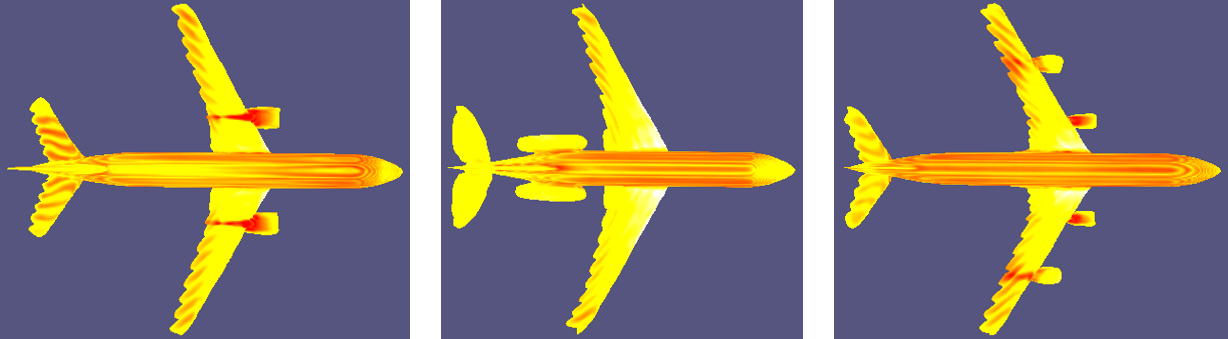}
}
\caption{The influence of $\gamma$ when $\lambda$ is set to $0$ in ISSN-CSL.}\label{fig:gamma1}
\vspace{-0.5em}
\end{figure}

We further explore the influence of $\lambda$ and $\gamma$ in ISSN-CSL. When we set $\lambda=0$, the influence of $\gamma$ is shown in Figure \ref{fig:gamma1}. We find that when $\gamma=1e-1, 1e-2, 1e-3$, the saliency maps does not show the `saliency' with almost the same saliency scores on the object surfaces and look less smooth, while when $\gamma=1.0$, the saliency maps focus on the fuselages more than other parts and look much smoother.

\begin{figure}[htbp]
\centering
\subfloat[$\lambda=1e-3, \gamma=1e-2$]{
\includegraphics[width=0.90\linewidth]{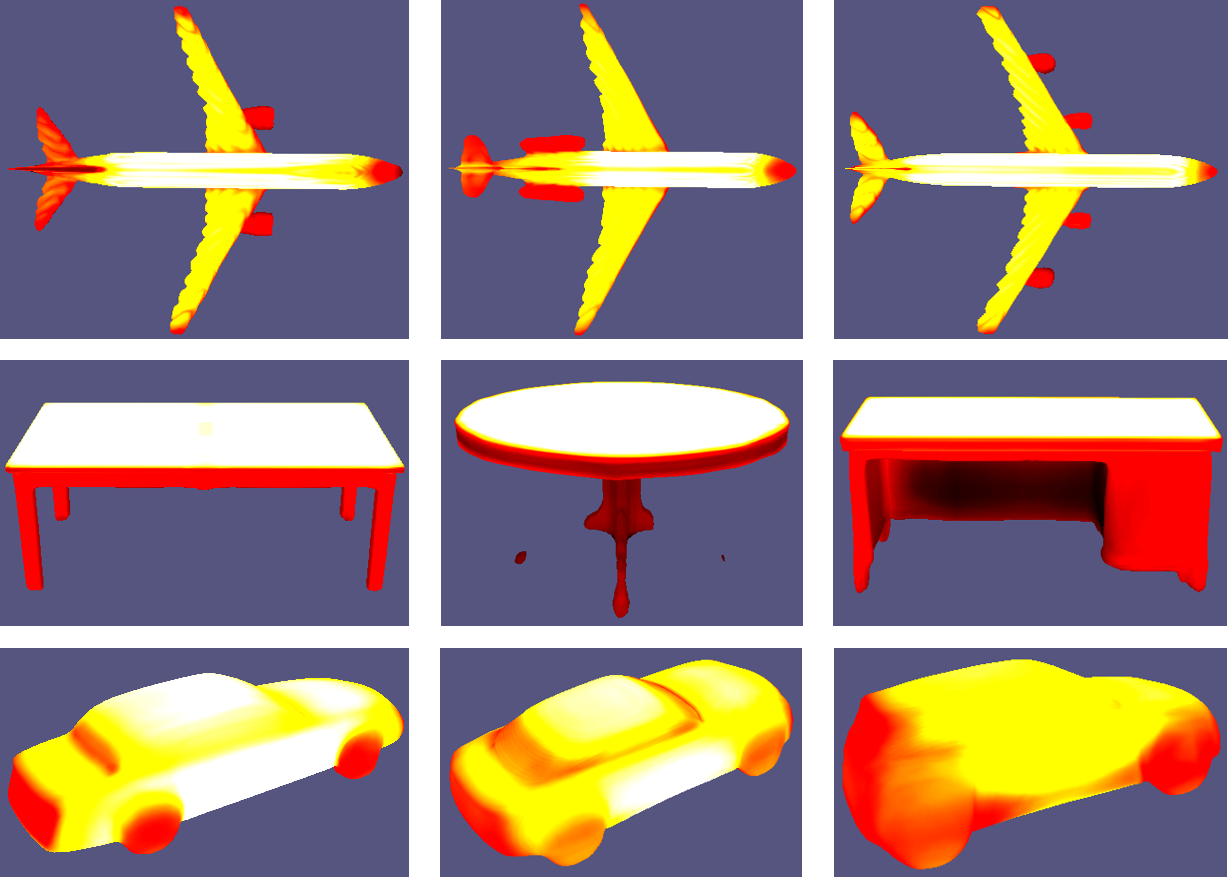}
}\\
\subfloat[$\lambda=1e-3, \gamma=1e-3$]{
\includegraphics[width=0.90\linewidth]{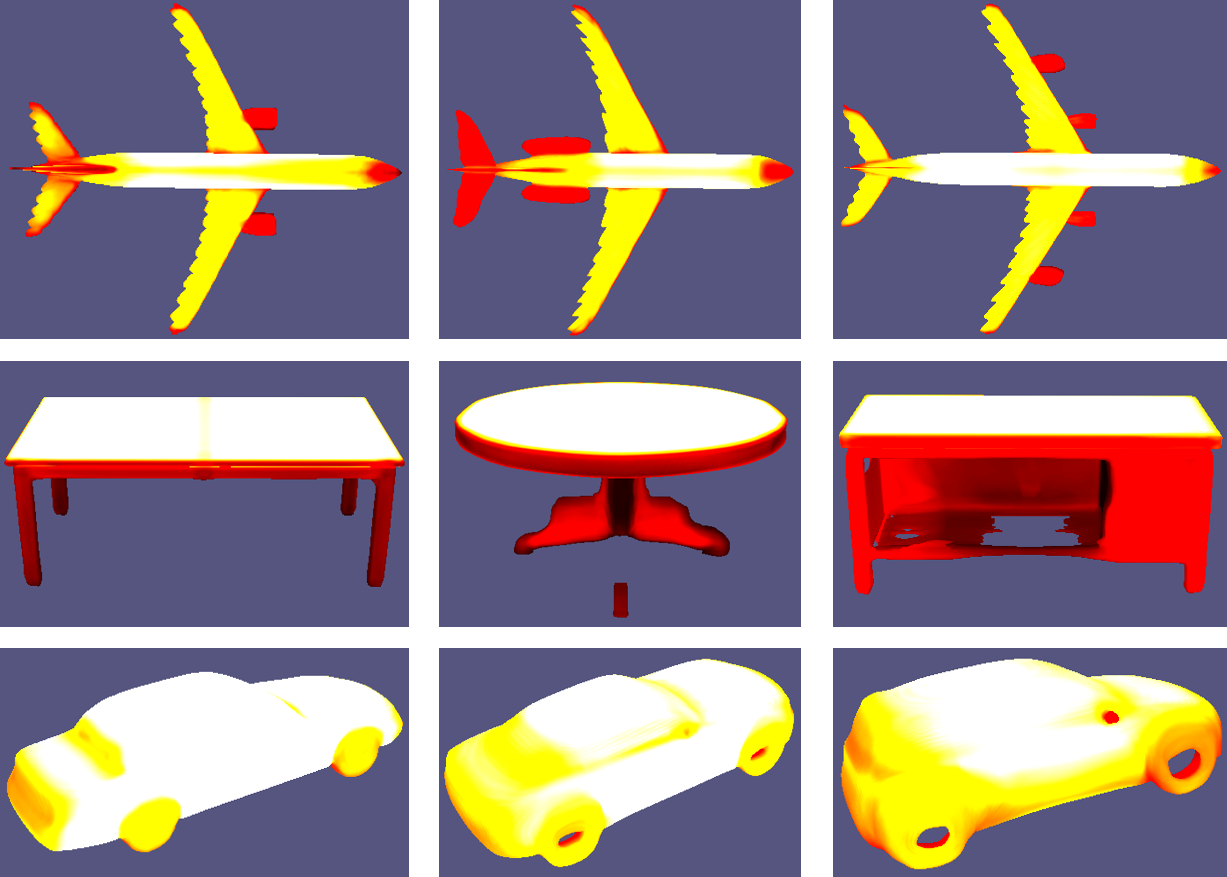}
}\\
\subfloat[$\lambda=1e-4, \gamma=1e-3$]{
\includegraphics[width=0.90\linewidth]{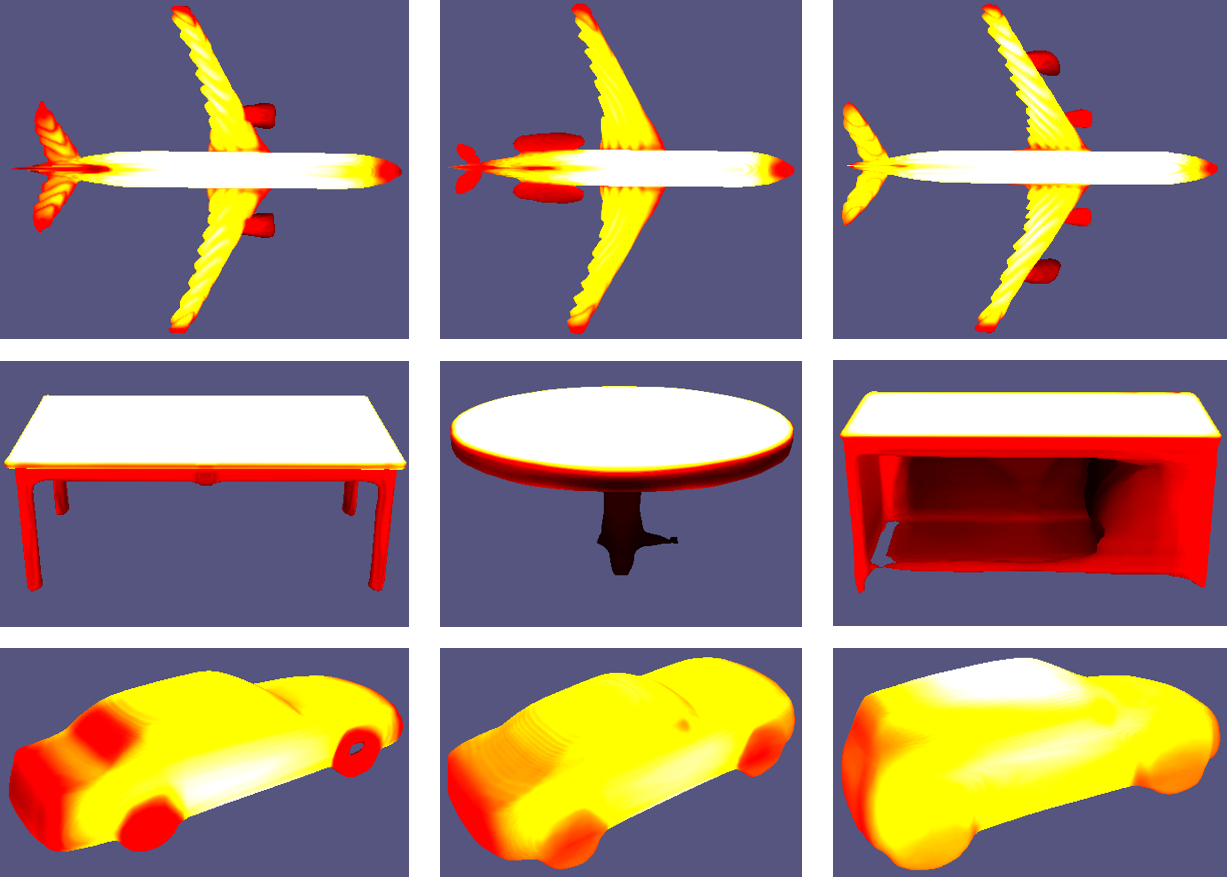}
}\\
\caption{The influence of $\lambda$ and $\gamma$. These models are from ShapeNet \cite{chang2015shapenet}.}\label{fig:gamma&lambda}
\vspace{-0.5em}
\end{figure}

When $\lambda>0$, we experiment that if the ratio of $\frac{\gamma}{\lambda}>=1000$, we can not reconstruct the object surfaces from the learned signed distance fields. In Section \ref{sup:issn}, we are aware that when $\lambda>=1e-2$, the learned saliency maps have high values all over the object surfaces, so we can not set $\gamma>=1.0$. Therefore, in the main paper, we choose $\lambda=1e-3$ and $\gamma=0.1$. Here, we present more examples when $\lambda$ and $\gamma$ are set to other values in Figure \ref{fig:gamma&lambda}. Corresponding saliency smoothness ratios and symmetry distances on ShapeNet \cite{chang2015shapenet} are shown in Table \ref{sup_tab:shapenet_smooth} and Table \ref{sup_tab:shapenet_sym}. 
Compared to our original setting in the main paper, the saliency maps of these three settings are comparably symmetric but less smooth.

\begin{table}[htbp]
    \centering
        \begin{tabular}{  l | c | c | c  }
             % \hline
            \toprule[1pt]
            ISSN-CSL         &  Mean   & Min     & Max    \\
            \hline
            $\lambda=1e-3, \gamma=1e-2$  &  0.052  &  0.006  &  0.224   \\
            $\lambda=1e-3, \gamma=1e-3$  &  0.059  &  0.010  &  0.247   \\
            $\lambda=1e-4, \gamma=1e-3$  &  0.056  &  0.006  &  0.242   \\
            \bottomrule[1pt]
        \end{tabular}
    \caption{Statistical SSR of objects on ShapeNet \cite{chang2015shapenet}.}
    \label{sup_tab:shapenet_smooth}
    \vspace{-0.5em}
\end{table}

\begin{table}[htbp]
    \centering
        \begin{tabular}{  l | c | c | c  }
             % \hline
            \toprule[1pt]
            ISSN-CSL         &  Mean   & Min     & Max    \\
            \hline
            $\lambda=1e-3, \gamma=1e-2$  &  0.073  &  0.008  &  0.493   \\
            $\lambda=1e-3, \gamma=1e-3$  &  0.070  &  0.018  &  0.492   \\
            $\lambda=1e-4, \gamma=1e-3$  &  0.085  &  0.009  &  0.496   \\
            \bottomrule[1pt]
        \end{tabular}
    \caption{Statistical symmetry distance of saliency map on ShapeNet \cite{chang2015shapenet}.}
    \label{sup_tab:shapenet_sym}
    \vspace{-0.5em}
\end{table}

%------------------------------------------------------------------------
\section{More Visualization of Saliency Maps} \label{sup:visual}
\begin{figure}[htbp]
\centering
\subfloat[ISSN ($\lambda=1e-3)$]{
\includegraphics[width=0.99\linewidth]{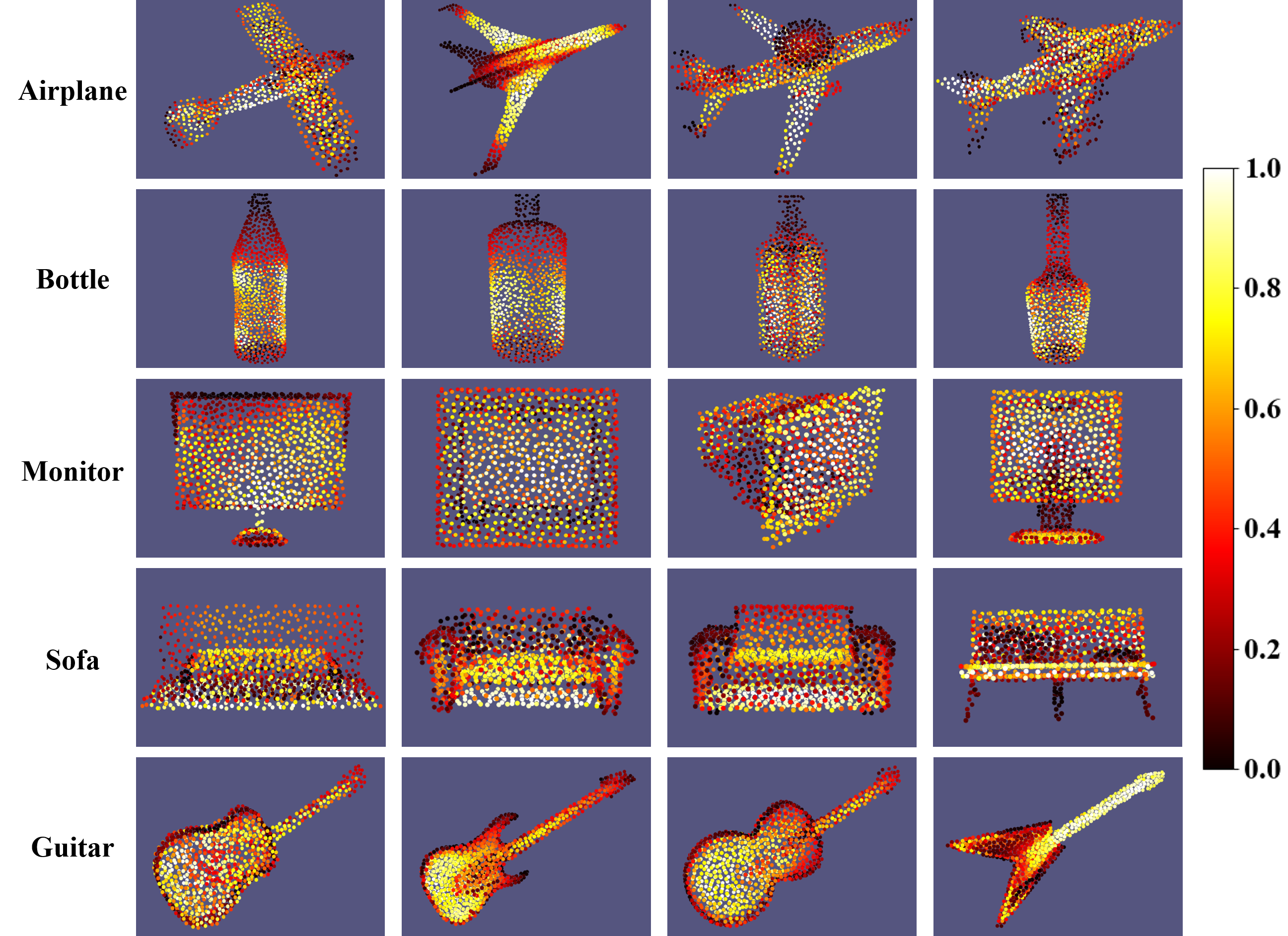}
}\\
\subfloat[ISSN-CSL ($\lambda=0, \gamma=1.0$)]{
\includegraphics[width=0.99\linewidth]{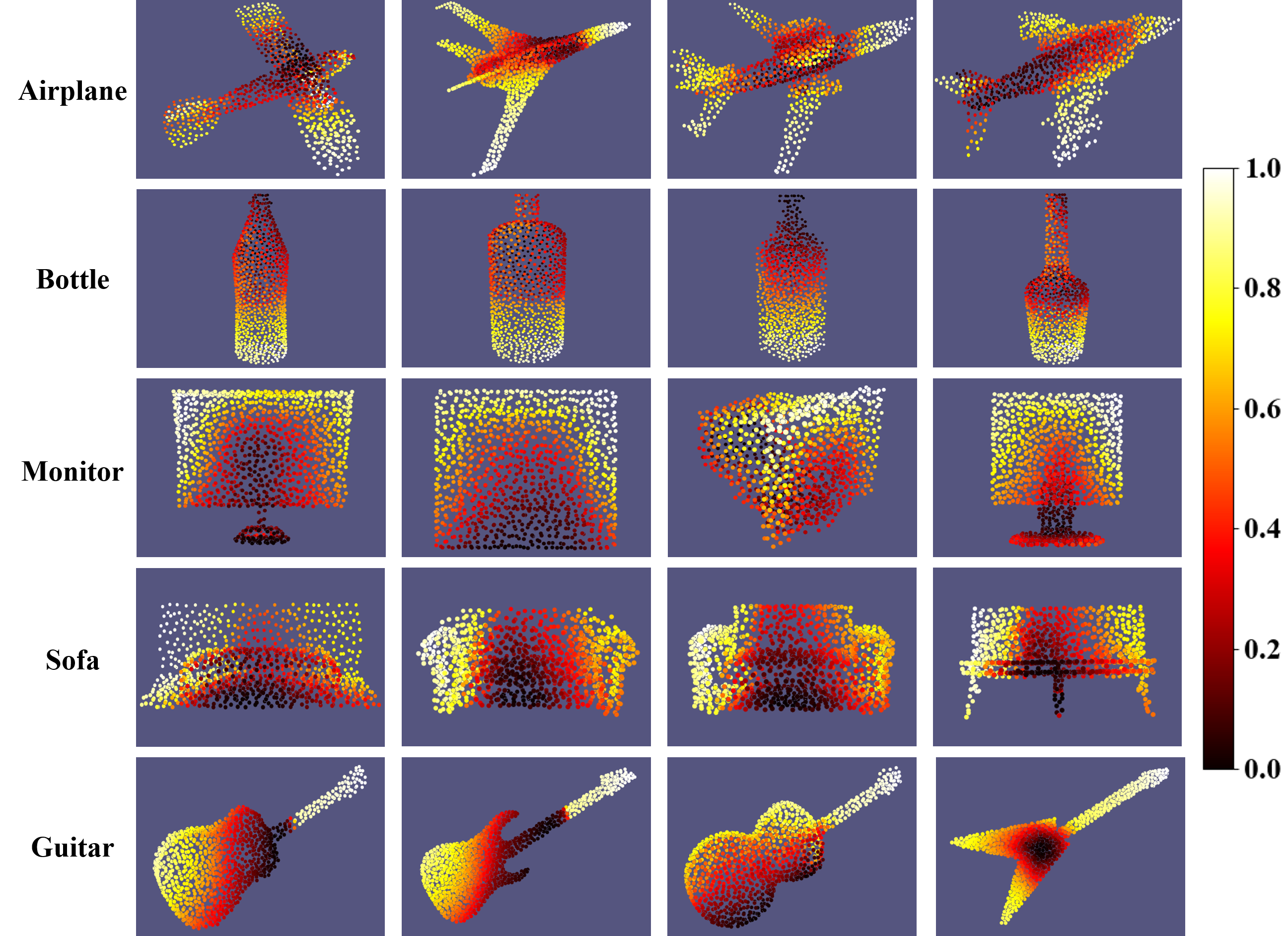}
}\\
\subfloat[ISSN-CSL ($\lambda=1e-3, \gamma=0.1$)]{
\includegraphics[width=0.99\linewidth]{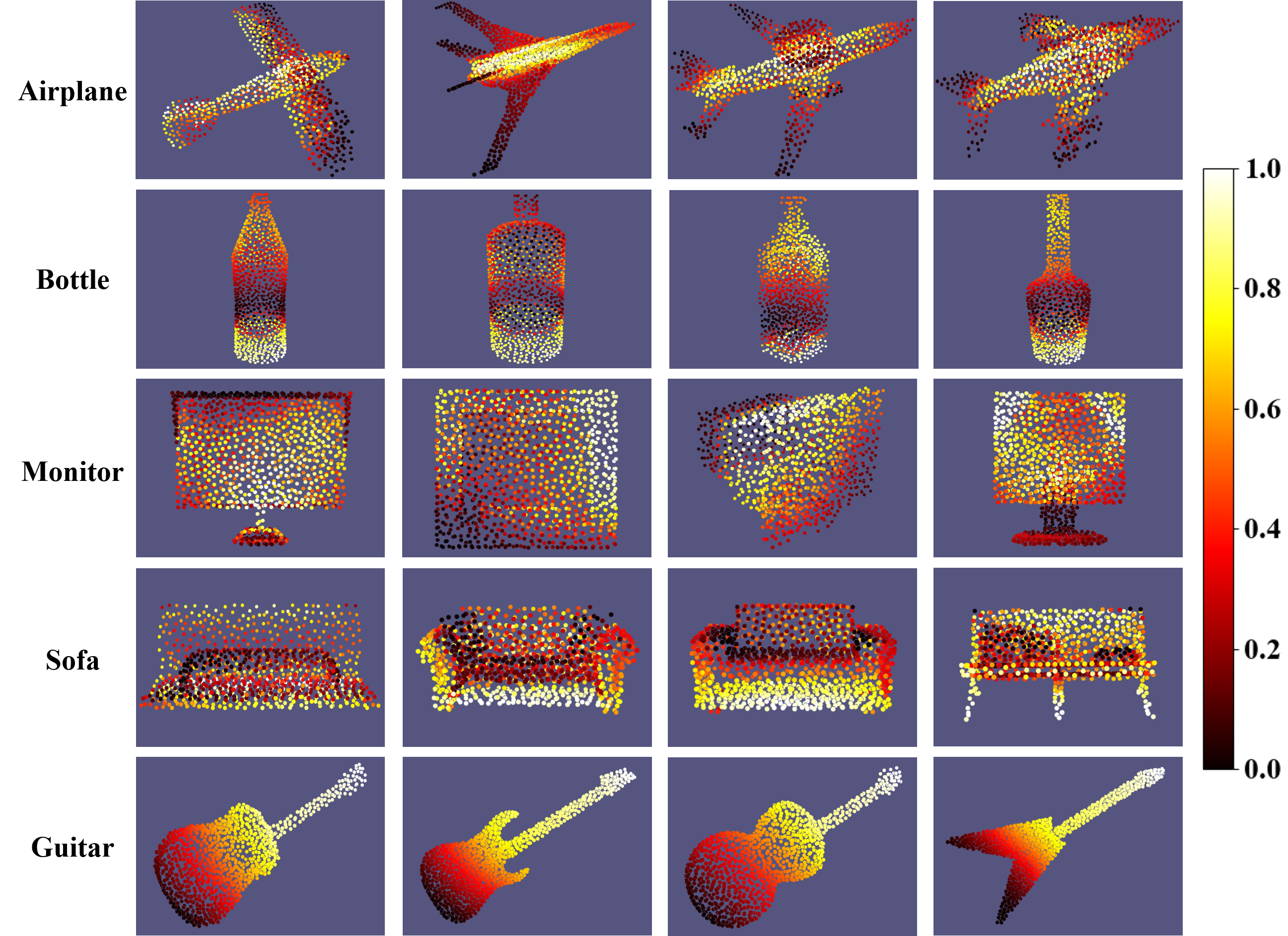}
}
\caption{Saliency maps on point clouds. These models are from ModelNet40 \cite{wu20153d}.}\label{sup_fig:pc_sal}
\vspace{-0.5em}
\end{figure}

We present more examples of saliency maps on point clouds estimated by ISSN and ISSN-CSL in Figure \ref{sup_fig:pc_sal}. We can see ISSN and ISSN-CSL can learn to generate continuous and smooth saliency map along the manifold of the object surface and maintain the symmetry property for those symmetric objects. Our proposed ISSN indeed capture the representative and common parts shared among the objects of the same category while ISSN-CSL can capture the discriminative parts betwwen different categories. 

%------------------------------------------------------------------------

\section{Training and Testing Curves on Saliency Points Classification} \label{sup:curves}

\begin{figure}[htbp]
\centering
\subfloat[Training curves]{
\includegraphics[width=0.99\linewidth]{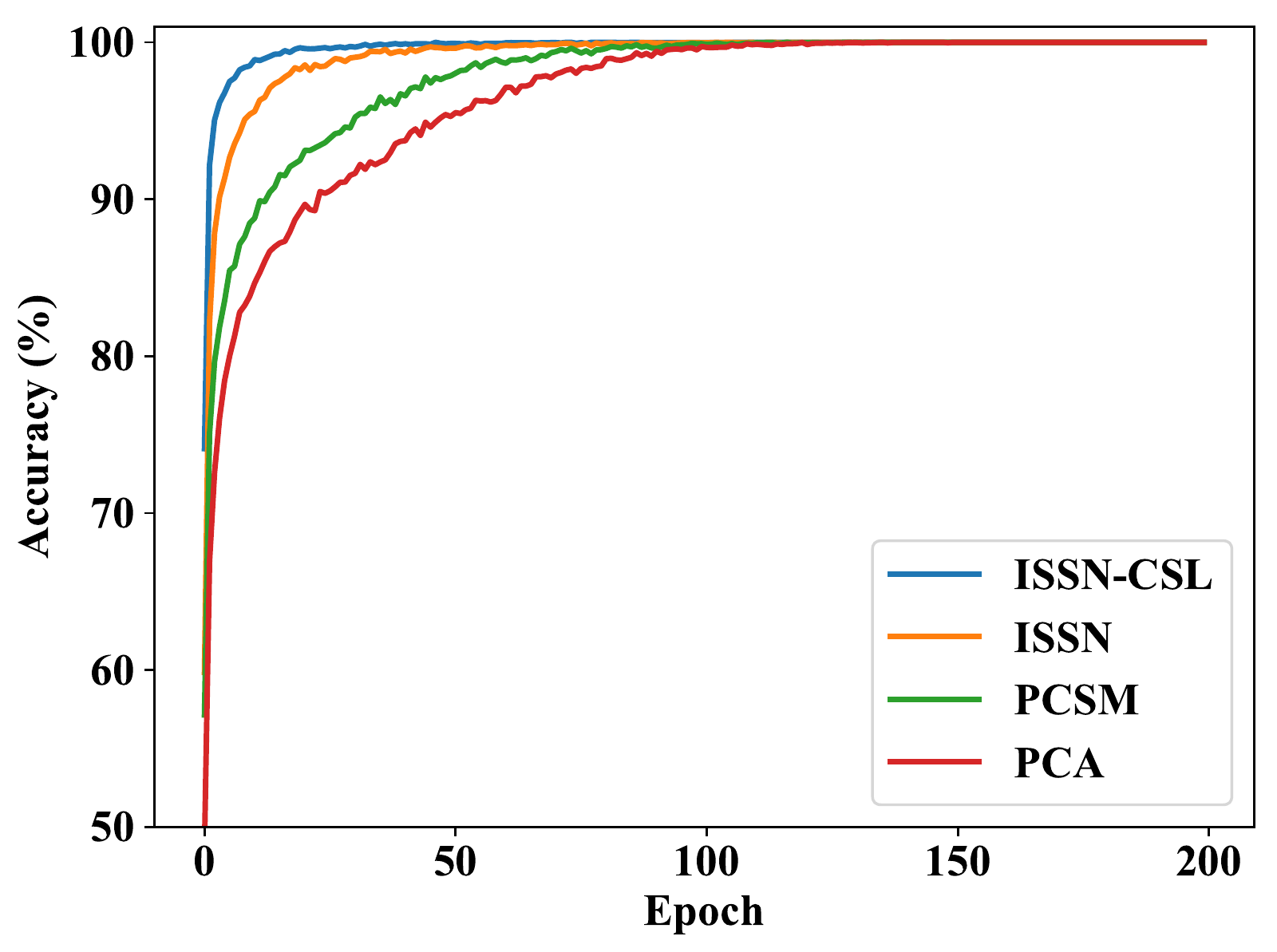}
}\\
\subfloat[Testing curves]{
\includegraphics[width=0.99\linewidth]{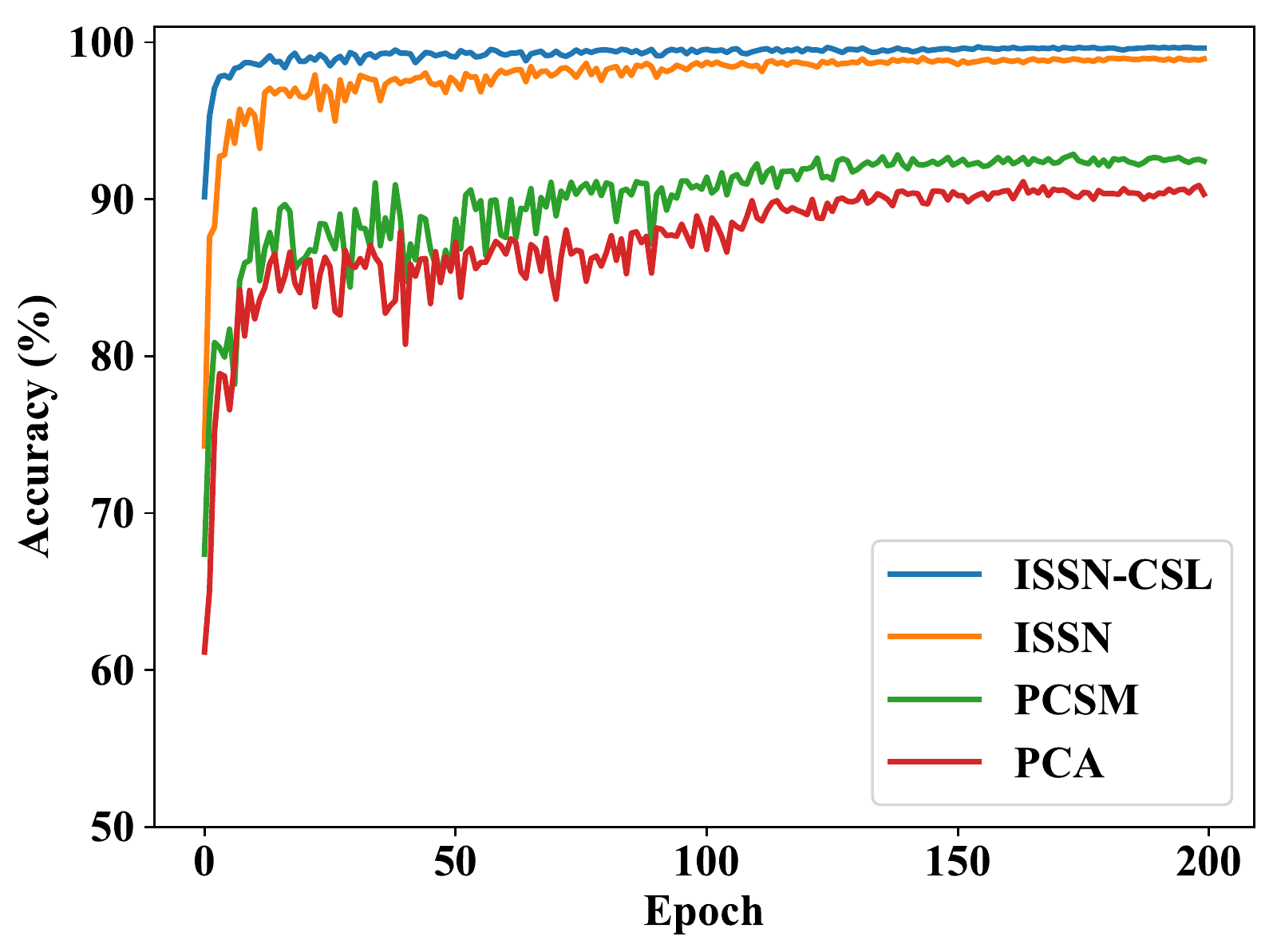}
}
\caption{Overall accuracy curves of training and testing of saliency points classification on ModelNet40 \cite{wu20153d} using DGCNN \cite{wang2019dynamic}. $512$ saliency points of each object are used for training and testing.}\label{fig:curves}
\vspace{-0.5em}
\end{figure}

We show the training and testing accuracy curves of saliency points classification on ModelNet40 \cite{wu20153d} using DGCNN \cite{wang2019dynamic} in Figure \ref{fig:curves}. We observe that ISSN and ISSN-CSL have better convergence rate compared to that of PCSM \cite{zheng2019pointcloud} and PCA \cite{bae2008method}. We can find that after $120$ epochs, all the methods become saturated on training set. However, ISSN and ISSN-CSL can achieve about $99\%$ accuracy on test set, while PCSM and PCA only reach about $93\%$ on test set, illustrating that the saliency points selected by ISSN and ISSN-CSL are more representative and the models trained on them generalize well from training set to test set, while the the saliency points selected by PCSM and PCA are less representative and the models trained on them are more prone to overfitting.

\typeout{get arXiv to do 4 passes: Label(s) may have changed. Rerun}

\end{document}